\documentclass[final,1p,times,twocolumn]{elsarticle}

\usepackage{amssymb}
\usepackage{amsmath}
\usepackage{natbib} 
\usepackage{url}
\usepackage{graphicx}%
\usepackage{multirow}
\usepackage{amsmath,amssymb,amsfonts}%
\usepackage{amsthm}%
\usepackage{mathrsfs}%
\usepackage[title]{appendix}%
\usepackage{xcolor}%
\usepackage{textcomp}%
\usepackage[numbers]{natbib}

\usepackage{manyfoot}%
\usepackage{booktabs}%
\usepackage{algorithm}%
\usepackage{algorithmicx}%
\usepackage{algpseudocode}%
\usepackage{listings}%

\journal{Image and Vision Computing}

\begin{document}
\title{YOLO Evolution: A Comprehensive Benchmark and Architectural Review of YOLOv12, YOLO11, and Their Previous Versions}
\author[1,2]{Nidhal Jegham}
\ead{jeghamnidhal7@gmail.com, nidhal.jegham@uri.edu}

\author[2]{Chan Young Koh}
\ead{ckoh04@uri.edu}

\author[2,3]{Marwan Abdelatti}
\ead{mabdelrazik@uri.edu, mabdelat@providence.edu}

\author[2]{Abdeltawab Hendawi\corref{cor1}}
\ead{hendawi@uri.edu}
\cortext[cor1]{Corresponding author}

\affiliation[1]{organization={Tunis Business School, University of Tunis}, 
            city={Tunis},
            country={Tunisia}}

\affiliation[2]{organization={Computer Science Department, University of Rhode Island}, 
            city={Kingston},
            state={RI},
            country={USA}}

\affiliation[3]{organization={Department of Computer Science, Providence College}, 
            city={Providence},
            state={RI},
            country={USA}}
\begin{abstract}
This study presents a comprehensive benchmark analysis of various YOLO (You Only Look Once) algorithms. It represents the first comprehensive experimental evaluation of YOLOv3 to the latest version, YOLOv12, on various object detection challenges. The challenges considered include varying object sizes, diverse aspect ratios, and small-sized objects of a single class, ensuring a comprehensive assessment across datasets with distinct challenges. To ensure a robust evaluation, we employ a comprehensive set of metrics, including Precision, Recall, Mean Average Precision (mAP), Processing Time, GFLOPs count, and Model Size. Our analysis highlights the distinctive strengths and limitations of each YOLO version. For example, YOLOv9 demonstrates substantial accuracy but struggles with detecting small objects and efficiency, whereas YOLOv10 exhibits relatively lower accuracy due to architectural choices that affect its performance in overlapping object detection but excels in speed and efficiency. Additionally, the YOLO11 family consistently shows superior performance maintaining a remarkable balance of accuracy and efficiency. However, YOLOv12 delivered underwhelming results, with its complex architecture introducing computational overhead without significant performance gains. These results provide critical insights for both industry and academia, facilitating the selection of the most suitable YOLO algorithm for diverse applications and guiding future enhancements.
\end{abstract}
\maketitle






\section{Introduction}
\label{sect:introduction}

Object detection is an essential component of computer vision systems, enabling automated identification and localization of objects within images or video frames \cite{liu2020deep}. Its applications span from autonomous driving and robotics \cite{feng2021review, treetrunk, s18041212, shoman2024enhancingtrafficsafetyparallel} to inventory management, video surveillance, and sports analysis \cite{babila2022object, 8457076, pedvehicle, sports}.

Over the years, object detection has developed significantly. Initially, traditional methods such as the Viola-Jones algorithm \cite{jones} and the Deformable Part-based Model (DPM) \cite{Felzenszwalb} used handcrafted features and were effective for applications such as face detection \cite{jones}, pedestrian detection \cite{pedestrian}, and video surveillance \cite{surveillance}. However, these methods had limitations in robustness and generalization. With the advancement of deep learning,  network-based methods have since become the primary approach. These methods are usually classified into two categories: one-stage and two-stage approaches.

One-stage methods such as RetinaNet \cite{lin2018focallossdenseobject} and SSD (Single Shot MultiBox Detector) \cite{Liu_2016} perform detection in a single pass, balancing speed and accuracy. In contrast, two-stage methods, such as Region-based Convolutional Neural Networks (R-CNN) \cite{girshick2014richfeaturehierarchiesaccurate}, generate region proposals and then perform classification, offering high precision but being computationally intensive.

YOLO (You Only Look Once) is a pioneering one-stage object detection framework that predicts bounding boxes and class probabilities in a single evaluation, enabling real-time performance. First introduced by Redmon et al. in 2015 \cite{redmon2016you}, YOLOv1 redefined object detection with its efficiency. YOLOv2 \cite{redmon2017yolo9000} improved upon this by integrating Darknet-19, batch normalization, and data augmentation techniques inspired by the VGG architecture \cite{simonyan2015deepconvolutionalnetworkslargescale}, leading to better generalization. YOLOv3 \cite{redmon2018yolov3} adopted Darknet-53, a deeper network with enhanced feature extraction, and incorporated a Feature Pyramid Network (FPN)-inspired design for multi-scale detection, significantly improving accuracy for objects of varying sizes. Subsequent iterations diversified under different development groups. YOLOv4 \cite{bochkovskiy2020yolov4} introduced Spatial Pyramid Pooling (SPP) for multi-scale feature fusion and the Path Aggregation Network (PAN) to refine feature integration. YOLOv5 \cite{ultralytics2021yolov5} marked a shift from the Darknet framework to PyTorch, increasing accessibility and efficiency through strided convolution layers and Spatial Pyramid Pooling Fast (SPPF) layers. YOLOv6 \cite{li2022yolov6} implemented RepVGG for simplified inference and CSPStackRep blocks for improved accuracy. YOLOv7 \cite{wang2023yolov7} introduced Extended Efficient Layer Aggregation Networks (E-ELAN) to optimize information flow and enhance detection performance.

More recent versions further pushed the boundaries of efficiency and accuracy. YOLOv8 \cite{sohan2024review}, released by Ultralytics, introduced scalable models tailored for various hardware constraints and expanded capabilities to tasks like semantic segmentation, pose estimation, and oriented bounding box (OBB) detection. YOLOv9 \cite{wang2024yolov9} introduced Programmable Gradient Information (PGI) for optimized gradient flow and Generalized Efficient Layer Aggregation Networks (GELAN) for enhanced feature fusion. YOLOv10 \cite{wang2024yolov10} eliminated Non-Maximum Suppression (NMS) using a dual assignment strategy, while also introducing lightweight classification heads and spatial-channel decoupled downsampling for increased efficiency. YOLOv11 \cite{yolo11_ultralytics} retained YOLOv8’s multi-task versatility while improving efficiency with the C3k2 block and introducing the C2PSA module for better spatial attention, particularly benefiting small and overlapping object detection. The latest iteration, YOLOv12, builds on prior advancements by incorporating an attention-centric approach with Area Attention (A2), which enhances feature aggregation while maintaining real-time performance. It also introduces Residual Efficient Layer Aggregation Networks (R-ELAN) to improve optimization stability and convergence \cite{tian2025yolov12attentioncentricrealtimeobject}. These innovations solidify YOLO as a leading framework in real-time object detection, continuously improving accuracy, speed, and adaptability across various tasks.

\begin{figure*}
    \centering
    \includegraphics[width=14cm]{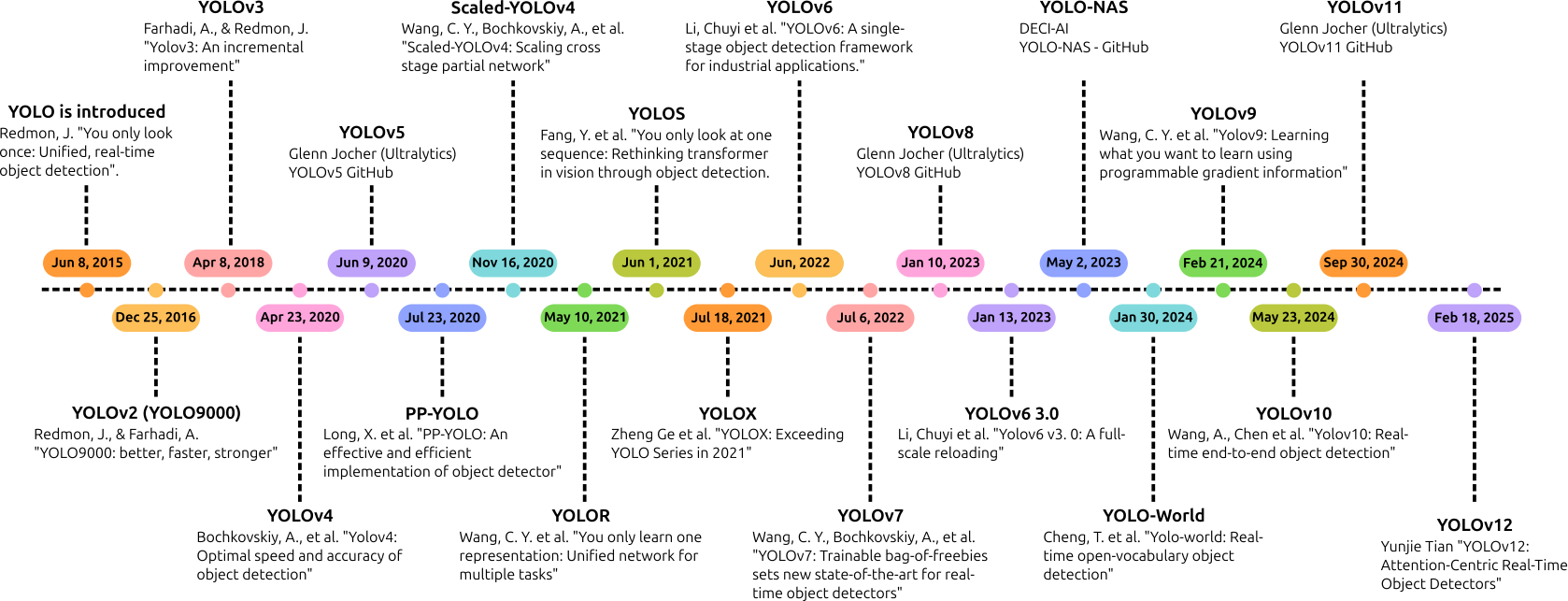}
    \vspace{-1em}
    \caption{Evolution of YOLO Algorithms throughout the years}
    \vspace{-1em}
    \label{fig:yoloevolution}
\end{figure*}

This object detection algorithm has undergone several developments, as seen in Figure \ref{fig:yoloevolution} achieving competitive results in terms of accuracy and speed, making it the preferred algorithm in various fields such as ADAS (Advanced Driver-Assist System) \cite{redmon2016you}, video surveillance \cite{mostafa2022yolo}, face detection \cite{nguyen2021yolo}, and many more \cite{garg2018deep}. For instance, YOLO plays a crucial role in the agriculture field by being implemented in numerous applications such as crop classification \cite{crops} \cite{crops2}, pest detection \cite{pestyolo}, automated farming \cite{wang2022mushroom, mirhaji2021fruit}, and virtual fencing \cite{vidya2021virtual}. Moreover, YOLO has been utilized on numerous occasions in the field of healthcare, such as cancer detection \cite{cancer2}, ulcer detection \cite{ulcer}, medicine classification \cite{mia2021real, patel2023identification}, and health protocols enforcement \cite{degadwala2021yolo}.

In recent years, Ultralytics has played a crucial role in the advancement of YOLO by maintaining, improving, and making these models more accessible \cite{rath2023yolov8}. Notably, Ultralytics has streamlined the process of fine-tuning and customizing YOLO models, a considerably more complex task in earlier iterations. The introduction of user-friendly interfaces, comprehensive documentation, and pre-built modules has greatly simplified essential tasks such as data augmentation, model training, and evaluation. Moreover, the development of scalable model versions allows users to select models tailored to specific resource constraints and application requirements, thereby facilitating more effective fine-tuning. The integration of advanced tools for hyperparameter tuning, automated learning rate scheduling, and model pruning has further refined the customization process. Continuous updates and robust community support helped YOLO models to be more accessible and adaptable for a wide range of applications.

This paper presents a comprehensive comparative analysis of the YOLO algorithm's evolution. It makes a significant contribution to the field by offering the first comprehensive evaluation of YOLOv12, the newest member of the YOLO family. By leveraging pre-trained models and fine-tuning them, we evaluate their performance across three diverse custom datasets, each with varying sizes and objectives. Consistent hyperparameters are applied to ensure a fair and unbiased comparison. The analysis focuses on numerous performance metrics, including speed, efficiency, accuracy, and computational complexity, as measured by GFLOPs count and model size. In addition, we explore the real-world applications of each YOLO version, highlighting their strengths and limitations across different use cases. Through this comparative study, we aim to provide valuable insights for researchers and practitioners, offering a deeper understanding of how these models can be effectively applied in various scenarios.

The rest of this paper is organized as follows: Section \ref{relatedwork} covers related work. Section \ref{sect:benchmark_setup} describes the datasets, models, experimental setup, hyperparameters, and evaluation metrics used. Section \ref{sect:benchmark} presents the experimental results, comparative analysis, and discussion, highlighting key discoveries, trends, architectural review, and real-life applications. Finally, Section \ref{sect:conclusion} concludes with insights drawn from the study.

\section{Related Work}\label{relatedwork}

The YOLO (You Only Look Once) algorithm is considered one of the most prominent object detection algorithms. It achieves state-of-the-art speed and accuracy, and its various applications have made it indispensable in numerous fields and industries. Numerous researchers have shown interest in this object detection algorithm by publishing papers reviewing its evolution, fine-tuning its models, and benchmarking its performance against other computer vision algorithms. This widespread interest underscores YOLO's important role in advancing the field of computer vision.

The paper \cite{fabio2024benchcloudvision} examines seven semantic segmentation and detection algorithms, including YOLOv8, for cloud segmentation from remote sensing imagery. It conducts a benchmark analysis to evaluate their architectural approaches and identify the most performing ones based on accuracy, speed, and potential applications. The research aims to produce machine learning algorithms with cloud segmentation using a few spectral bands, including RGB and RGBN-IR combinations.

Highlighting advancements in UAV-based object detection, the study \cite{khan2024visionary} introduces SFFEF-YOLO to tackle challenges posed by varying target scales and small object prevalence in aerial imagery. By replacing the large prediction head with a tiny one, integrating a Fine-Grained Information Extraction Module (FIEM), and refining multi-scale fusion with the Multi-Scale Feature Fusion Module (MFFM), the model achieves an increase of 9.9\% on VisDrone2019-DET and 3.6\% on UAVDT in terms of mAP50 over YOLOv8. However, using an older YOLO version may limit its full potential. Incorporating newer YOLO models such as YOLO11 could further enhance accuracy, efficiency, and generalizability, a gap this paper addresses through broader benchmarks.

The authors of \cite{hussain2024yolov1} review the evolution of the YOLO variants from version 1 to version 8, examining their internal architecture, key innovations, and benchmarked performance metrics. The paper highlights the models' applications across domains like autonomous driving and healthcare and proposes incorporating federated learning to improve privacy, adaptability, and generalization in collaborative training. The review, however, limits its focus to mAP (mean Average Precision) for accuracy evaluation, neglecting other key metrics such as Recall and Precision. Additionally, it considers FPS (frames per second) as the sole measure of computational efficiency, excluding the impact of preprocessing, inference, postprocessing times, GFLOPs, and size.

The paper \cite{diwan2023object} thoroughly analyzes single-stage object detectors, particularly YOLOs from YOLOv1 to YOLOv4, with updates to their architecture, performance metrics, and regression formulation. Additionally, it provides an overview of the comparison between two-stage and single-stage object detectors and applications utilizing two-stage detectors. However, not including newer YOLO models limits its comprehensiveness, leaving a gap in understanding the advancements and improvements introduced in more recent versions.

The authors of \cite{sapkota2024yolov10} explore the evolution of the YOLO algorithms from version 1 to 10, highlighting their impact on automotive safety, healthcare, industrial manufacturing, surveillance, and agriculture. The paper highlights incremental technological advances and challenges in each version, indicating a potential integration with multimodal, context-aware, and General Artificial Intelligence systems for future AI-driven applications. However, the paper does not include a benchmarking study or a comparative analysis of the YOLO models, leaving out performance comparisons across the versions.

The study \cite{KHAN2024105195} presents a lightweight rotational object detection algorithm to address challenges in remote sensing and surveillance, particularly variations in object size and orientation. By integrating an angle prediction branch and the Circular Smooth Label (CSL) method, along with a Channel and Spatial Attention (CSA) module, the model enhances feature extraction and detection accuracy. Achieving 57.86 mAP50 on the DOTA v2 dataset while maintaining a low computational footprint, it demonstrates efficiency for real-time deployment. However, its reliance on YOLOv5 limits potential performance gains. However, maintaining the backbone architecture of YOLOv5 can lead to limitations in terms of both computational efficiency and accuracy. Therefore, incorporating newer YOLO architectures could further improve the scope of this paper, 

The authors in the work in \cite{kang2023real} analyze the YOLO algorithm, focusing on its development and performance. They conduct a comparative analysis of the different versions of YOLO till the 8th version, highlighting the algorithm's potential to provide insights into image and video recognition and addressing its issues and limitations. The paper focuses exclusively on the mAP metric, overlooking other accuracy measures such as Precision and Recall. Additionally, it neglects speed and efficiency metrics, limiting the scope of the comparative study. The paper also omits evaluating the most recent models, YOLOv9, YOLOv10, and YOLO11.

This paper makes several key contributions: (i) It pioneers a comprehensive comparison of YOLOv12 against its predecessors across their scaled variants from nano- to extra-large; (ii) It provides deep insights into the architectural advancements of YOLO by analyzing key structural developments across versions. (iii) It evaluates YOLO models using three diverse datasets, reflecting various object properties and real-world applications, including Smart Cities, Satellite Imaging, and Wildlife Conservation. (iv) The performance evaluation extends beyond mAP and FPS, incorporating critical metrics such as Precision, Recall, Preprocessing Time, Inference Time, Postprocessing Time, GFLOPs, and model size. (v) These comprehensive metrics provide valuable insights for selecting the most optimal YOLO models, benefiting industry professionals and academics. (vi) It offers specific use case recommendations, identifying the most suitable YOLO models for different scenarios and environments, such as resource-constrained deployments, real-time applications, and the detection of small or overlapping objects.

\section{Benchmark Setup}\label{sect:benchmark_setup}

\subsection{Datasets}
This study conducts in-depth benchmark research and assesses the YOLO algorithms provided by the Ultralytics library. The main goal is to provide a thorough and comparative analysis of these models and explain their strengths, deficiencies, and possible applications.

This paper is made possible using several publicly accessible datasets on Kaggle and Roboflow. The selection of the datasets is based on the increasing implementation of the YOLO algorithms in the fields of Autonomous driving \cite{redmon2016you, driving1, driving2, driving3}, satellite imagery \cite{sat1, sat2, sat3}, and wildlife conservation \cite{wild1, wild2, wild3}. Moreover, each picked dataset presents unique difficulties and situations for object detection with varying image sizes and number of observations alongside the number of classes to ensure a comprehensive evaluation.  

\subsubsection{Traffic Signs Dataset}
\begin{figure*}[h]
    \centering
    \includegraphics[width=14cm]{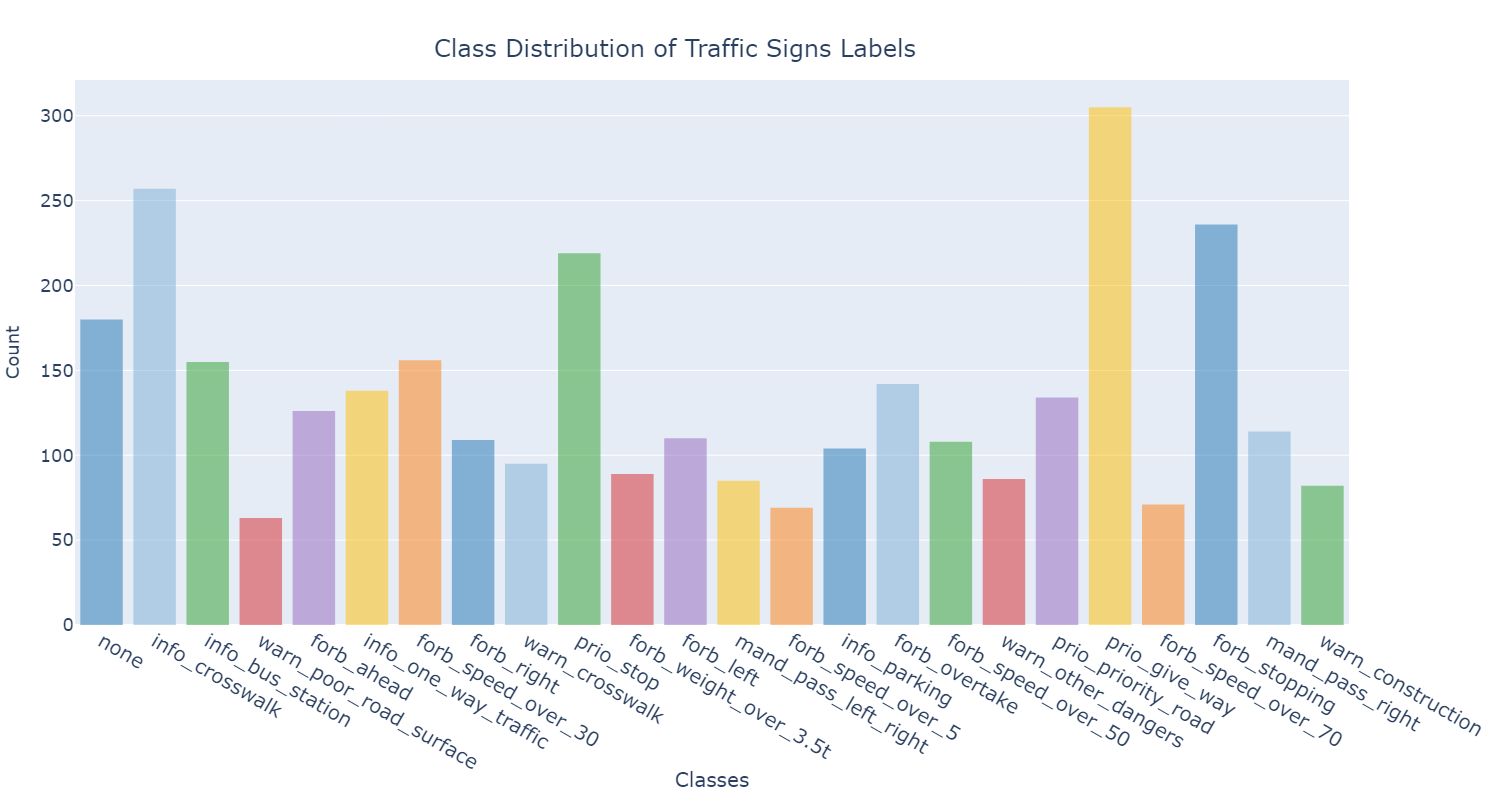}
    \caption{Classes Distribution of the Traffic Signs Dataset}
    \label{fig: distribution signs}
\end{figure*}

The Traffic Signs dataset by Radu Oprea is an open-source dataset on Kaggle containing around 55 classes across 3253 training and 1128 validation images of traffic signs in various sizes and environments \cite{traffic-signs-detection-europe_dataset}. All images in the dataset are initially sized 640$\times$640, with no labels for False Positives detection. To balance the classes, undersampling techniques were applied, as shown in Figure \ref{fig: distribution signs}. After preprocessing the dataset by removing classes with less than 50 observation counts, 24 classes remained, with 3233 images split into 70\% training, 20\% validation, and 10\% testing, with no data augmentation techniques applied. This dataset is vital for applications in autonomous driving, traffic management, road safety, and intelligent transportation systems. Additionally, it presents challenges due to the varying sizes of target objects and the similarities in patterns across different classes, which complicate the detection process.

\subsubsection{Africa Wild Life Dataset}
\begin{figure}[htp]
    \centering
    \includegraphics[width=8cm]{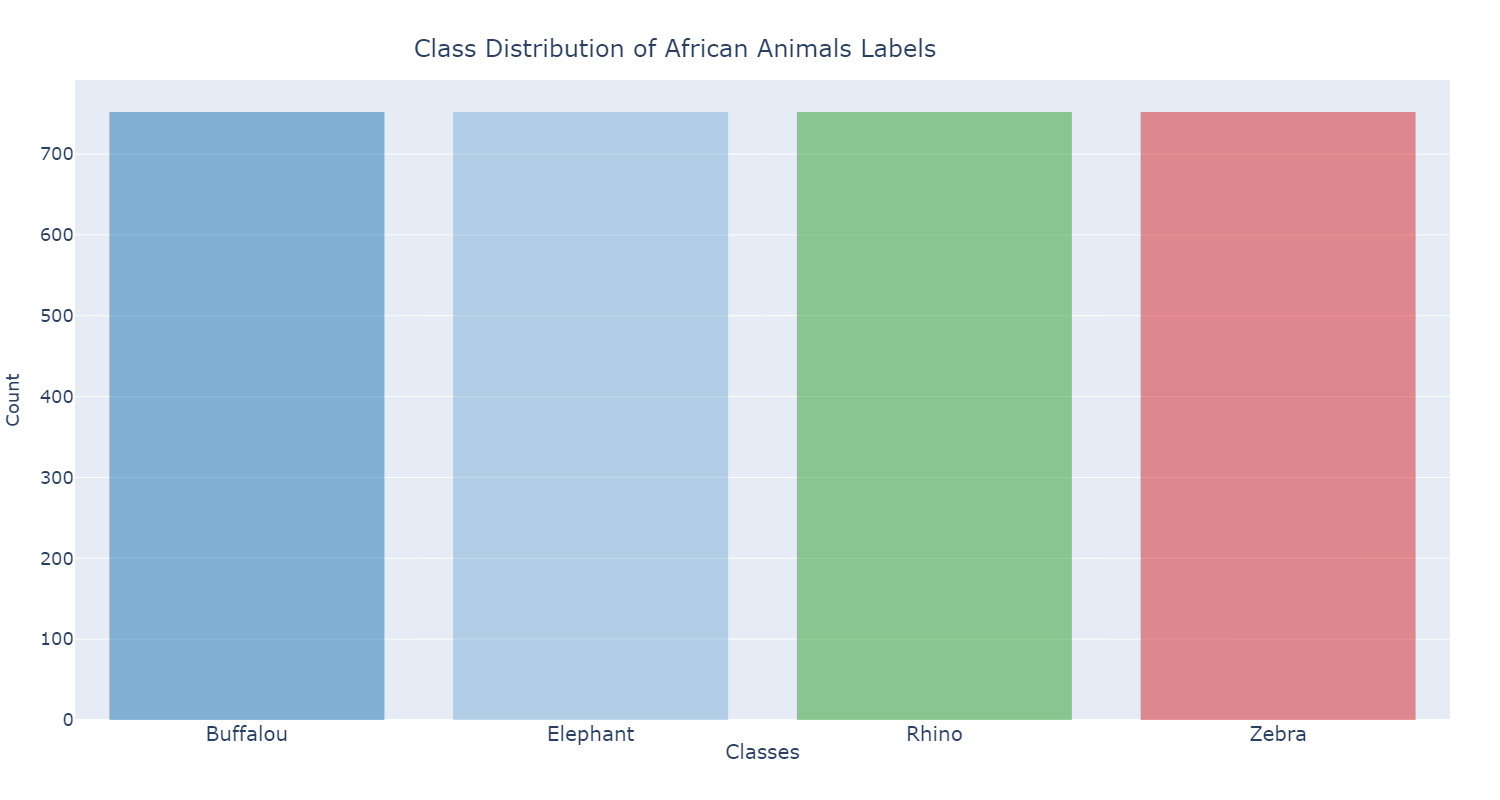}
    \caption{Classes Distribution of the Africa Wildlife Dataset}
    \label{fig: distribution africa}
\end{figure}

The Africa Wildlife dataset is an open-source Kaggle dataset by Bianca Ferreira, designed for real-time animal detection in nature reserves \cite{africa-wild-life_dataset}. It features four common African animal classes: Buffalo, elephant, rhino, and zebra. Each class is represented by at least 376 images, which were collected via Google image searches without any data augmentation techniques applied. The dataset is split into 70\% training, 20\% validation, and 10\% testing, with all images manually labeled in the YOLO format, as shown in Figure \ref{fig: distribution africa}. The dataset presents challenges such as varying aspect ratios, with each image containing at least one instance of the specified animal class and potentially multiple occurrences of other classes. Furthermore, the overlap of target objects increases the difficulty of detection. This dataset is crucial for applications in wildlife conservation, anti-poaching efforts, biodiversity monitoring, and ecological research.

\subsubsection{Ships/Vessels Dataset}
The Ships/Vessels dataset is an extensive open-source collection containing approximately 13.5k images, collected by Siddharth Sah from numerous Roboflow datasets, curated explicitly for ship detection \cite{ships-vessels_dataset}. Each image has been manually annotated with bounding boxes in the YOLO format, ensuring precise and efficient detection of ships. This dataset features a single class, "ship," and is divided into 70\% training, 20\% validation, and 10\% testing. Notably, no data augmentation techniques were applied during the experiment, which ensures that the model's performance is evaluated on the raw, unaltered dataset. However, the relatively small size of the target objects, varying aspect ratios, and their varying rotations pose challenges for detection, particularly for the YOLO algorithm, which often struggles with small object detection and objects with varying orientations. The dataset is essential for various practical applications such as maritime safety, fisheries management, marine pollution monitoring, defense, maritime security, and more.
\subsection{Models}

\subsubsection{Ultralytics vs. Original YOLO}
In this subsection, we will conduct a comparative analysis between the models provided by Ultralytics and their original counterparts on the Traffic Signs dataset provided by Radu Oprea \cite{traffic-signs-detection-europe_dataset} using the same hyperparameters in Table \ref{tab:parameters}.  The objective is to highlight the differences between Ultralytics models and the original versions, which justifies the exclusion of YOLOv4 \cite{bochkovskiy2020yolov4}, YOLOv6 \cite{li2022yolov6}, and YOLOv7 \cite{wang2023yolov7} from this paper due to the lack of support for these models by Ultralytics. This analysis will demonstrate why focusing exclusively on Ultralytics-supported models ensures a fair and consistent benchmark evaluation.

\paragraph{\textbf{Ultralytics Supported Models and Tasks:}}
Ultralytics library provides researchers and programmers various YOLO models for inference, validation, training, and export. Based on the results of Table \ref{tab:yolo_ultra}, we notice that Ultralytics does not support YOLOv1, YOLOv2, YOLOv4, and YOLOv7. Concerning YOLOv6, the library only supports the configuration *.yaml files without the pre-trained *.pt models.

\begin{table}[]
\centering
\caption{Ultralytics-supported library tasks and models}
\begin{tabular}{lccc}
\hline
\textbf{\begin{tabular}[c]{@{}l@{}}YOLO \\ Version\end{tabular}} & \textbf{Inference} & \textbf{Validation} & \textbf{Training} \\ \hline
YOLOv1                                                           & No                 & No                  & No                \\
YOLOv2                                                           & No                 & No                  & No                \\
\textbf{YOLOv3u}                                                 & Yes                & Yes                 & Yes               \\
YOLOv4                                                           & No                 & No                  & No                \\
\textbf{YOLOv5u}                                                 & Yes                & Yes                 & Yes               \\
YOLOv6                                                           & Yes                & Yes                 & Yes               \\
YOLOv7                                                           & No                 & No                  & No                \\
\textbf{YOLOv8}                                                  & Yes                & Yes                 & Yes               \\
\textbf{YOLOv9}                                                  & Yes                & Yes                 & Yes               \\
\textbf{YOLOv10}                                                 & Yes                & Yes                 & Yes               \\
\textbf{YOLO11}                                                  & Yes                & Yes                 & Yes               \\
\textbf{YOLOv12}                                                 & Yes                & Yes                 & Yes              \\
\bottomrule
\end{tabular}
\footnotetext{Boldface text shows available pre-trained models}
\label{tab:yolo_ultra} 
\end{table}
\begin{figure}
    \centering
    \includegraphics[width=5cm] {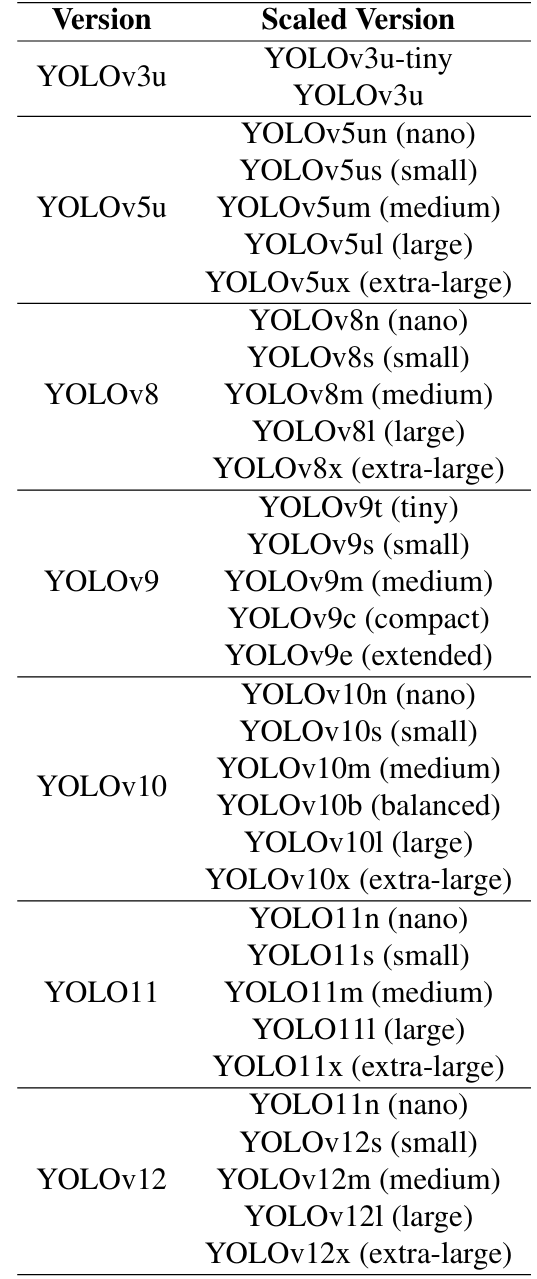}
    \caption{YOLO versions and scaled versions}
    \label{fig: YOLO Families}
\end{figure}
\paragraph{\textbf{Performance Comparison of Ultralytics and Original Models:}}
Based on the results of our comparative analysis of the Ultralytics models and their original counterparts on the Traffic Signs dataset presented in Table \ref{tab:yolo_comparison}, we observe significant discrepancies between the performance of the Ultralytics models and their original counterparts. Notably, Ultralytics' versions of YOLOv5n (nano) and YOLOv3 demonstrate superior performance, underscoring the enhancements and optimizations implemented by Ultralytics. Conversely, the original YOLOv9c (compact) slightly outperforms its Ultralytics version, potentially due to the lack of extensive optimization for this newer model by Ultralytics. These observations highlight that the Ultralytics models have undergone substantial modifications, making a direct comparison with the original versions inequitable. Consequently, the noticeable performance discrepancy between the two models, including the original and Ultralytics models in the same benchmarking study, would not provide a fair or accurate assessment. Therefore, this paper will focus exclusively on the Ultralytics-supported versions to ensure consistent and fair benchmarks.

In total, 33 models from 7 different YOLO versions were trained on three different datasets, as seen in Figure \ref{fig: YOLO Families}. Each YOLO version includes several scaled models (e.g., YOLOv5u, YOLOv5un, YOLOv5us, and YOLOv5ux), with the suffixes denoting model size and complexity, such as "n" for nano, "s" for small, "m" for medium, "l" for large, "x" for extra-large, "t" for tiny, "c" for compact, "b" for balanced, and "e" for extended. These models offer a variety of trade-offs between accuracy
and inference speed, as discussed in the following sections.

\begin{table}[]
    \centering
\caption{Ultralytics and original YOLO performance comparison}
\centering
\begin{tabular}{cccc}
\hline
\textbf{Version} & \textbf{Source} & \textbf{mAP50} & \textbf{mAP50-95} \\ \hline
YOLOv9c          & Ultralytics     & 0.845          & 0.748             \\
                 & Github          & 0.881          & 0.786             \\
YOLOv5n          & Ultralytics     & 0.756          & 0.663             \\
                 & Github          & 0.429          & 0.367             \\
YOLOv3           & Ultralytics     & 0.766          & 0.67              \\
                 & Github          & 0.562          & 0.471             \\ \hline
\end{tabular}
\label{tab:yolo_comparison} 

\end{table}

\subsubsection{YOLOv3u}
\begin{figure}
    \centering
    \includegraphics[scale=0.26, trim={1cm 0 0 0}, clip] {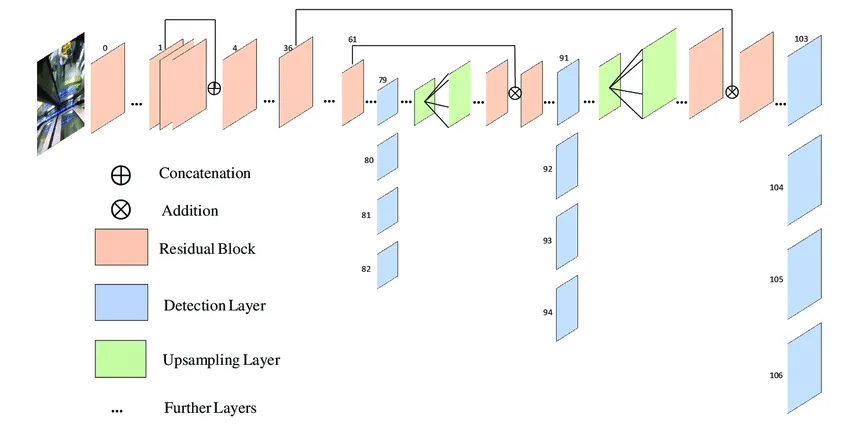}
    \caption{YOLOv3 architecture showcasing the residual blocks and the upsampling layers \cite{yolov3_benchmark}}
    \label{fig: YOLOv3 Architecture}
\end{figure}

YOLOv3 enhances localization and detection efficiency, particularly for small objects, using the Darknet-53 framework, which offers double the speed of ResNet-152 \cite{redmon2018yolov3}. It integrates Feature Pyramid Network (FPN) elements, including residual blocks, skip connections, and up-sampling, to improve multi-scale detection, as illustrated in Figure \ref{fig: YOLOv3 Architecture}. The model generates feature maps at three scales (down-sampling at factors of 32, 16, and 8) for detecting objects of varying sizes. However, YOLOv3 struggles with medium and large objects, leading Ultralytics to introduce YOLOv3u, which adopts an anchor-free detection method, later used in YOLOv8, improving both accuracy and speed.

\subsubsection{YOLOv5u}
\begin{figure}
    \centering
    \includegraphics[scale=0.7]{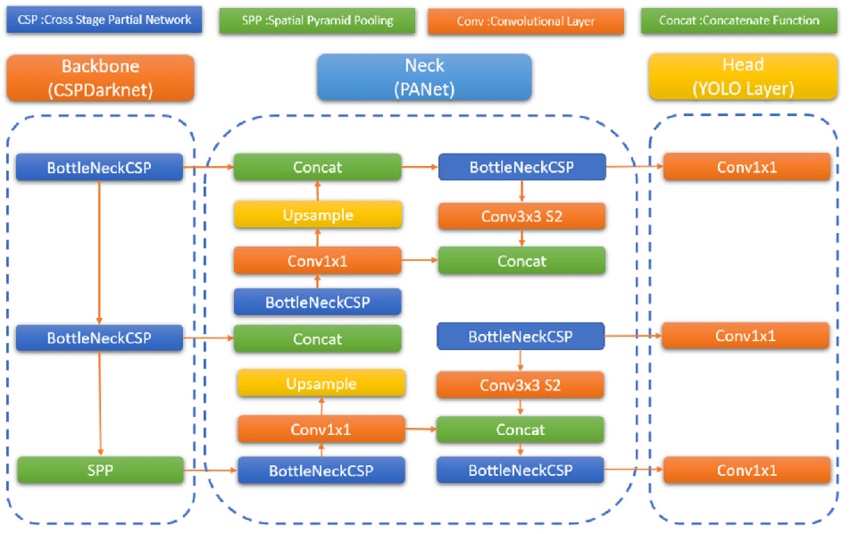}
    \caption{Detailed architecture of YOLOv5 including the CSPDarknet Backbone, PANet Neck, and YOLO Layer Head \cite{yolov5diagram}}
    \label{fig: YOLOv5 Architecture}
\end{figure}
YOLOv5, developed by Glenn Jocher, transitions from Darknet to PyTorch, incorporating CSPDarknet as its backbone for improved feature extraction and reduced computational cost \cite{ultralytics2021yolov5} \cite{yolov5}. It features Cross-Stage Partial (CSP) connections, a strided convolution layer, and the Spatial Pyramid Pooling Fast (SPPF) module, which enhances multi-scale feature representation, as illustrated in Figure~\ref{fig: YOLOv5 Architecture}. YOLOv5 also integrates data augmentation techniques (Mosaic, copy-paste, MixUp, HSV augmentation) to improve robustness. Available in five scaled variants, it continues to evolve with YOLOv5u, which introduces anchor-free detection for enhanced accuracy and efficiency, particularly on complex objects of varying sizes.

\subsubsection{YOLOv8}

\begin{figure}
    \centering
    \includegraphics[scale=0.25]{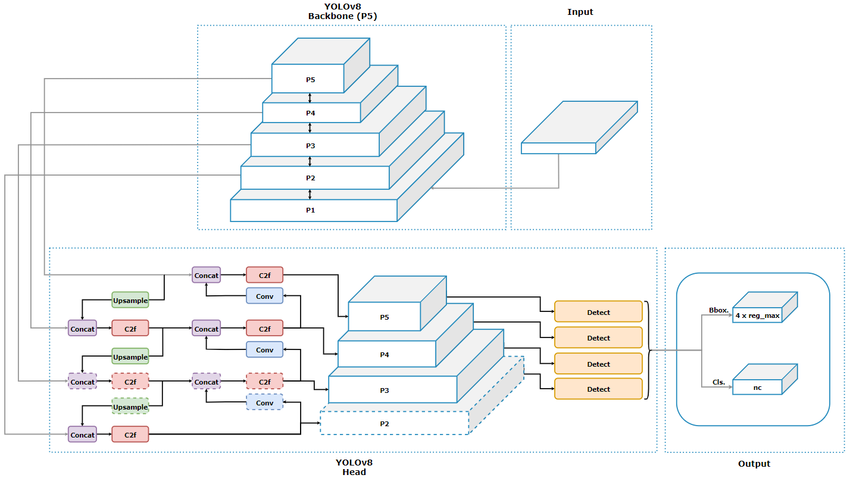}
    \caption{Detailed architecture of YOLOv8 showcasing the backbone's multiple convolutional layers to extract hierarchical features, the Feature Pyramid Network (FPN) enhances detection at different scales. The network head performs final predictions, incorporating convolutional blocks and upsample blocks to refine features \cite{yolov8diagram}}
    \label{fig: YOLOv8 Architecture}
\end{figure}

Ultralytics has introduced YOLOv8, a significant evolution in the YOLO series, with five scaled versions \cite{sohan2024review} \cite{yolov8_ultralytics}. Alongside object detection, YOLOv8 also provides various applications such as image classification, pose estimation, instance segmentation, and OBB. Key features include a backbone similar to YOLOv5, with adjustments in the CSPLayer, now known as the C2f module, which combines high-level features with contextual information for enhanced detection accuracy highlighted in Figure \ref{fig: YOLOv8 Architecture}. YOLOv8 also introduces a semantic segmentation model called YOLOv8-Seg, which combines a CSPDarknet53 feature extractor with a C2F module, achieving state-of-the-art results in object detection and semantic segmentation benchmarks while maintaining high efficiency. 

\subsubsection{YOLOv9}

\begin{figure}
    \centering
    \includegraphics[scale=0.4, trim={1cm 0 0 0}, clip]{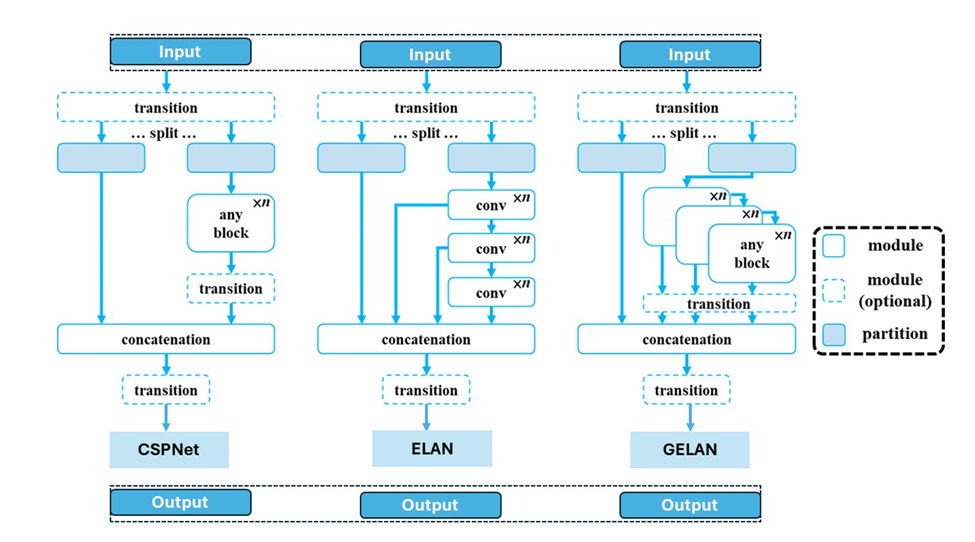}
    \caption{YOLOv9 architecture featuring CSPNet, ELAN, and GELAN modules. CSPNet optimizes gradient flow and reduces computational complexity via feature map partitioning. ELAN enhances learning efficiency by linearly aggregating features, and GELAN extends this concept by integrating features from various depths and pathways \cite{wang2024yolov9}}
    \label{fig: YOLOv9 Architecture}
\end{figure}
YOLOv9, developed by Chien-Yao Wang, I-Hau Yeh, and Hong-Yuan Mark Liao, enhances object detection using the Information Bottleneck Principle and Reversible Functions, ensuring efficient gradient propagation and improved model convergence \cite{wang2024yolov9}. Reversible functions enable lossless information retention, benefiting lightweight models prone to under-parameterization. A key advancement, Programmable Gradient Information (PGI), dynamically adjusts gradient flow during training, prioritizing informative gradients to prevent feature loss. Additionally, YOLOv9 integrates Gradient Enhanced Lightweight Architecture Network (GELAN), optimizing computational pathways for better parameter utilization, high-speed inference, and accurate detection, as illustrated in Figure~\ref{fig: YOLOv9 Architecture}. With five scalable versions, YOLOv9 balances efficiency and feature retention, making it highly adaptable for various real-world applications.

\subsubsection{YOLOv10}
\begin{figure}
    \centering
    \includegraphics[scale=0.2]{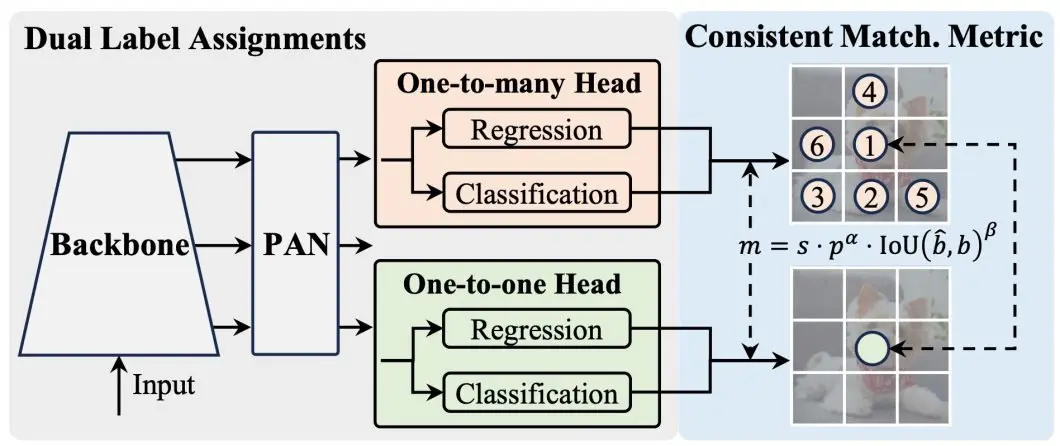}
    \caption{YOLOv10 architecture showcasing the dual label assignment strategy for improving accuracy. The PAN layer enhances feature representation alongside one-to-many head for regression and classification tasks and one-to-one head for precise localization \cite{wang2024yolov10}}
    \label{fig: YOLOv10 Architecture}
\end{figure}

YOLOv10, developed by Tsinghua University researchers, introduces key innovations, enhancing gradient flow and computational efficiency through an improved Cross Stage Partial Network (CSPNet) backbone \cite{wang2024yolov10}. The architecture, illustrated in Figure~\ref{fig: YOLOv10 Architecture}, consists of three main components: the backbone, the neck, and the detection head. The Path Aggregation Network (PAN) neck improves multi-scale feature fusion, allowing for better detection of objects at different sizes. The One-to-Many Head generates multiple predictions per object during training, enhancing learning accuracy.

For inference, YOLOv10 replaces Non-Maximum Suppression (NMS) with a One-to-One Head, producing a single best prediction per object, reducing latency and post-processing time. Additionally, NMS-Free Training leverages dual assignments, optimizing both speed and accuracy. Other enhancements include lightweight classification heads, spatial-channel decoupled downsampling, large-kernel convolutions, and partial self-attention modules, all improving efficiency without significantly increasing computational costs. YOLOv10 is available in five scaled versions, from nano to extra-large, catering to diverse use cases.

\subsubsection{YOLO11}
\begin{figure}[htp]
    \centering
    \includegraphics[width=8cm]{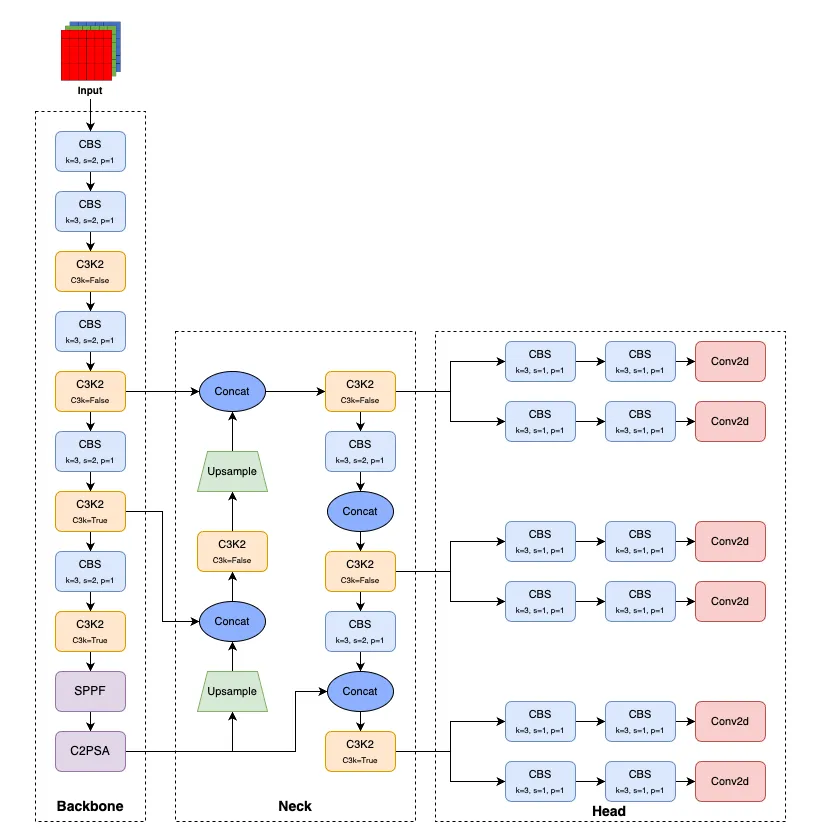}
    \caption{YOLO11 architecture showcasing the new C3k2 blocks and the C2PSA module \cite{yolo11_ultralytics} \cite{paulyolo112024}}
    \label{fig: YOLO11 Architecture}
\end{figure}
YOLO11 \cite{yolo11_ultralytics} is one of the latest additions to the YOLO series developed by Ultralytics, building upon the developments of its predecessors, especially YOLOv8. This iteration offers five scaled models from nano to extra large, catering to various applications. Like YOLOv8, YOLO11 includes numerous applications such as object detection, instance segmentation, image classification, pose estimation, and OBB.

Key improvements in YOLO11 include the introduction of the  Cross-Stage Partial with Self-Attention (C2PSA) module, as seen in Figure \ref{fig: YOLO11 Architecture}, which combines the benefits of cross-stage partial networks with self-attention mechanisms. This enables the model to capture contextual information more effectively across multiple layers, improving object detection accuracy, especially for small and colluded objects. Additionally, in YOLO11, the C2f block has been replaced by C3k2, a custom implementation of the CSP Bottleneck that uses two convolutions, unlike YOLOv8's use of one large convolution. This block uses a smaller kernel, retaining accuracy while improving efficiency and speed.

\subsubsection{YOLOv12}  
\begin{figure}[htp]
    \centering
    \includegraphics[width=8cm]{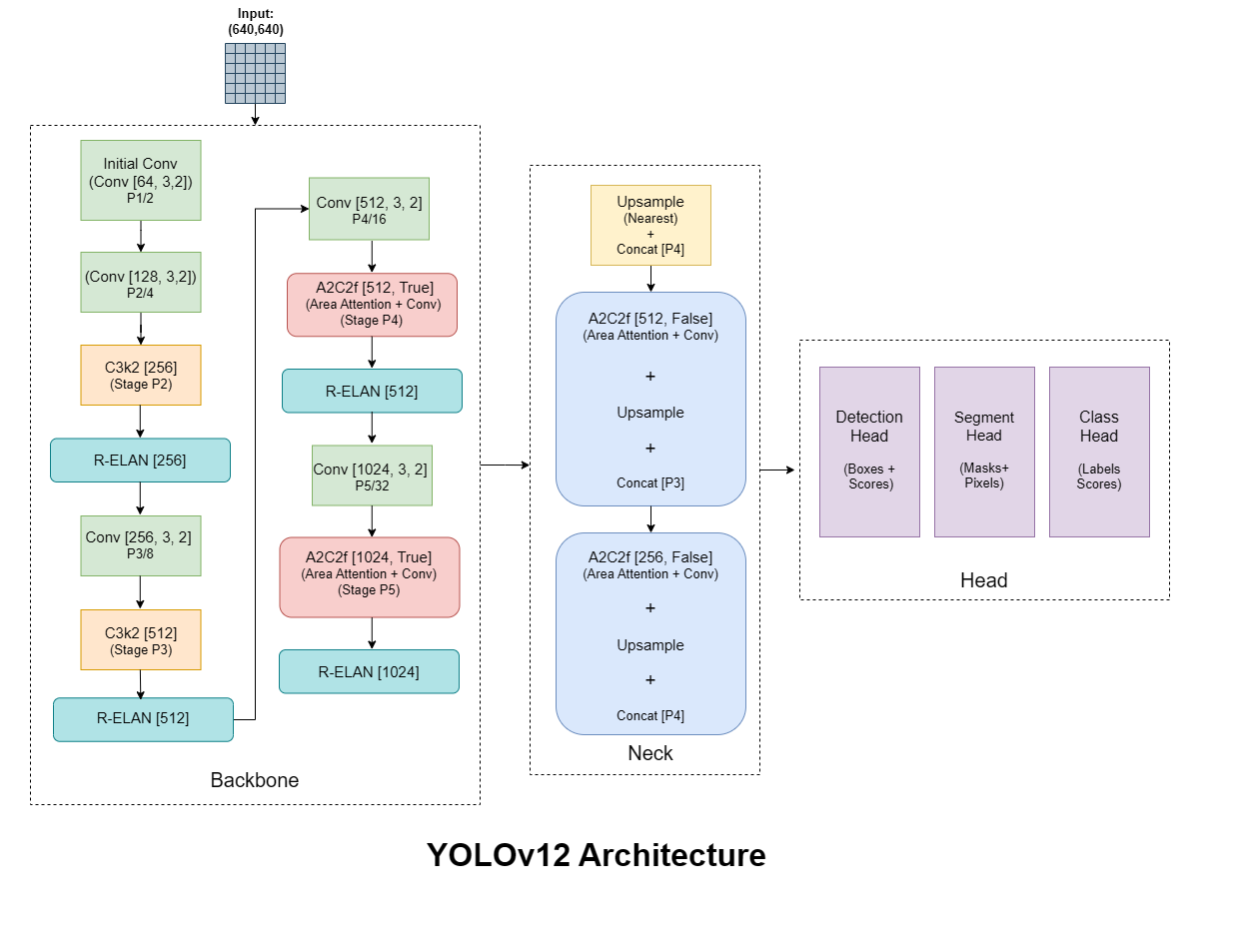}
    \caption{YOLOv12 architecture showcasing the new Area Attention (A2) module and Residual Efficient Layer Aggregation Networks (R-ELAN)}
    \label{fig: YOLOv12 Architecture}
\end{figure}
YOLOv12 \cite{tian2025yolov12attentioncentricrealtimeobject} is the latest evolution in the YOLO series, introducing an attention-centric design that significantly enhances both speed and accuracy. It continues the trend of offering five scalable models (Nano to Extra Large), making it adaptable for a wide range of applications such as object detection, instance segmentation, and OBB. As seen in Figure \ref{fig: YOLOv12 Architecture}, YOLOv12 introduces the Area Attention (A2) module, which maintains a large receptive field while drastically reducing computational complexity, allowing the model to enhance speed without compromising accuracy. Additionally, it features Residual Efficient Layer Aggregation Networks (R-ELAN), which improve training stability and model convergence through block-level residual design and optimized feature aggregation.

Moreover, YOLOv12 uses FlashAttention to minimize memory access overhead, closing the speed gap with CNNs. It adjusts the MLP ratio from 4 to 1.2, enhancing runtime efficiency, and removes positional encodings for a cleaner, faster model without losing detection accuracy. YOLOv12 also reduces complexity by using a single R-ELAN block in the last stage of the backbone instead of stacking three attention/CNN blocks. It replaces linear layers with convolutional layers and batch normalization, maximizing computational efficiency and achieving state-of-the-art latency-accuracy trade-offs.

\subsection{Hardware and Software Setup}
The experiments were conducted with Python 3.12, Ubuntu 22.04, CUDA 12.4, and cuDNN 8.9.7 for GPU acceleration. Ultralytics 8.2.55 was used for model training, while WandB 0.17.4 was utilized for tracking experiments. To ensure a fair comparison, similar hyperparameters were used across all models, as outlined in Table \ref{tab:parameters}. The experiments were carried out using 2 NVIDIA RTX 4090 GPUs, each with 16,384 CUDA cores.

\begin{table}
\centering
\caption{Table of parameters}
\centering
\begin{tabular}{lc}
\hline
\textbf{Parameter}             & \textbf{Value} \\ \hline
Epochs                         & 100            \\
Optimizer                      & AdamW          \\
Batch Size                     & 16             \\
Image Size                     & (640, 640)     \\
Initial \& Final Learning rate & (0.0001, 0.01) \\
Dropout rate                   & 0.15           \\
Augmentation Techniques        & None           \\
Data Split                     & (70, 20, 10)   \\ \hline
\end{tabular}
\label{tab:parameters}
\end{table}

\subsection{Metrics}

This study evaluates YOLO models based on accuracy, computational efficiency, and size. Accuracy metrics include Precision, Recall, mAP50, and mAP50-95. Precision measures the ratio of correctly predicted observations to total predictions, indicating False Positives, while Recall highlights False Negatives by comparing correct predictions to actual observations \cite{padilla2020survey}. mAP50 calculates Mean Average Precision at an IoU threshold of 0.50, whereas mAP50-95 extends this across thresholds from 0.50 to 0.95 with a 0.05 step size \cite{10028728}.

Computational efficiency is assessed using Preprocessing Time, Inference Time, and Postprocessing Time. Preprocessing Time refers to data preparation, Inference Time measures prediction generation, and Postprocessing Time converts raw outputs into final results. Additionally, GFLOPs indicate computational power, while model size reflects storage requirements and parameter count. These metrics enable a comprehensive performance comparison, ensuring a robust benchmark for real-world applications.

\section{Benchmark Results}\label{sect:benchmark}
\subsection{Results}
\subsubsection{Traffic Signs Dataset}

\begin{figure*}[htp]
    \centering
    \includegraphics[width=14cm]{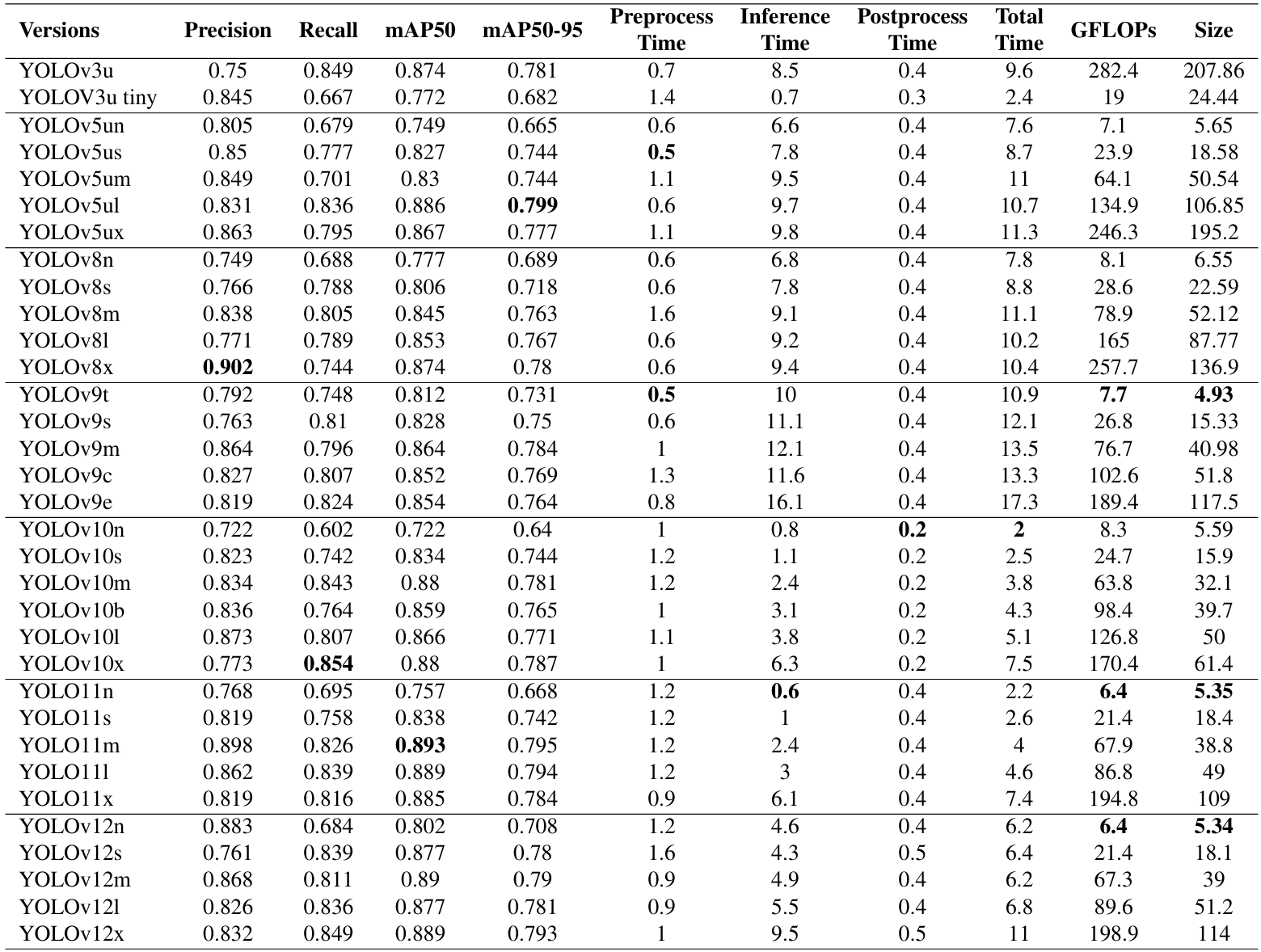}
    \caption{Evaluation results for the traffic signs dataset.}
    \label{fig: traffic signs table}
\end{figure*}

Figure \ref{fig: traffic signs table} presents a comparative analysis of the YOLO algorithms' performance on the Traffic Signs dataset, evaluated based on accuracy, computational efficiency, and model size. The Traffic Signs dataset is a medium-sized dataset with varied object sizes, making it favorable for benchmarking. 

\begin{figure*}[htp]
    \centering
    \includegraphics[width=14cm, trim={0 0 0 1.7cm}, clip]{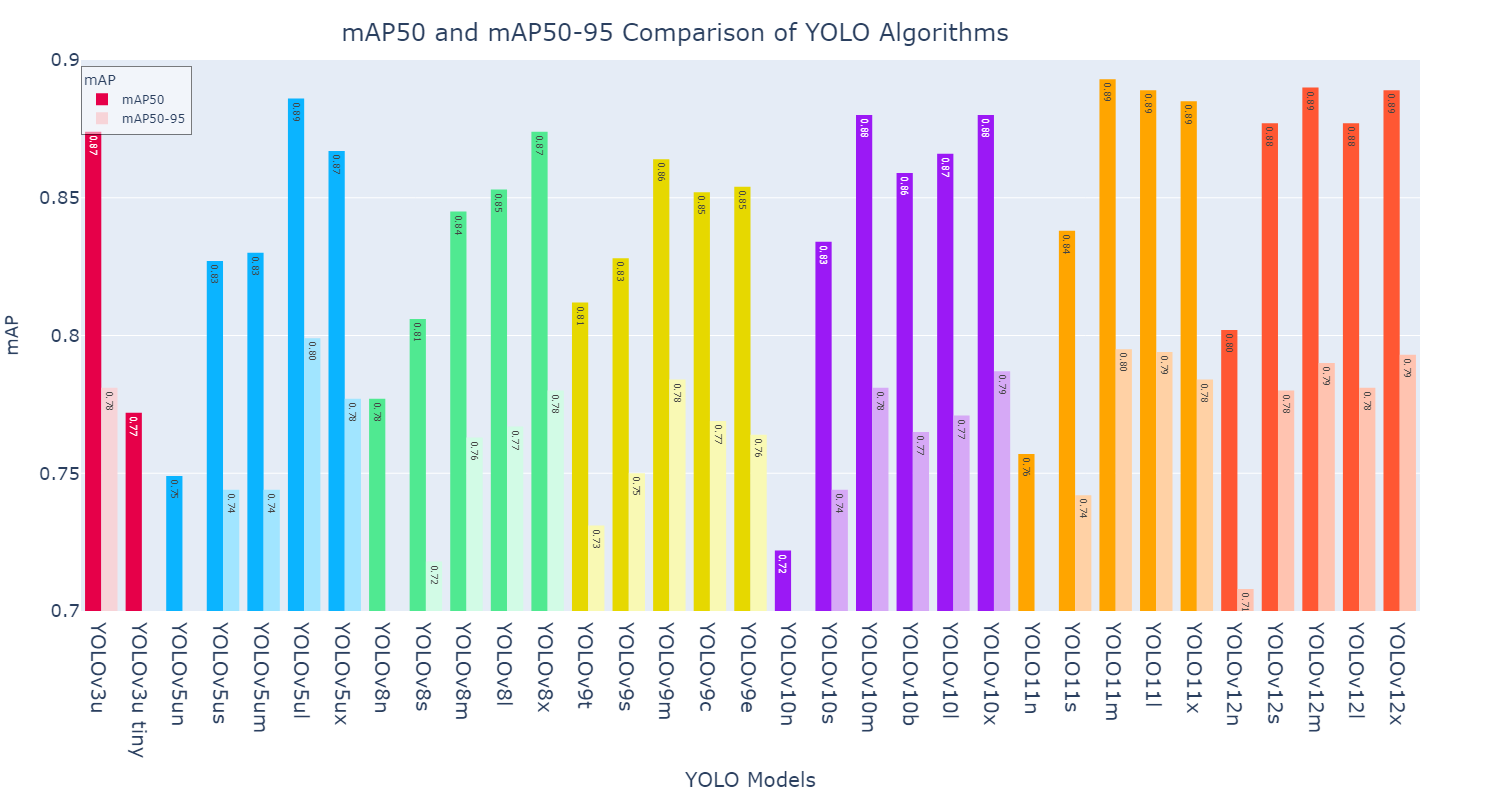}
    \caption{mAP50 and mAP50-95 YOLO results on traffic signs dataset. Each model is represented by two bars: the left bar shows the mAP50 score, while the right bar represents the mAP50-95 score}
    \label{fig: map signs}
\end{figure*}

\paragraph{\textbf{Accuracy:}}
 As illustrated in Figure \ref{fig: map signs}, YOLOv5ul demonstrates the highest accuracy, achieving a mAP50 of 0.866 and a mAP50-95 of 0.799. This is followed by YOLO11m with a mAP50-95 of 0.795 and YOLO11l with a mAP50-95 of 0.794. In contrast, YOLOv10n exhibits the lowest precision, with a mAP50 of 0.722 and a mAP50-95 of 0.64, closely followed by YOLOv5un with a mAP50-95 of 0.665.

\begin{figure*}[htp]
    \centering
    \includegraphics[width=14cm, trim={0 0 0 2cm}, clip]{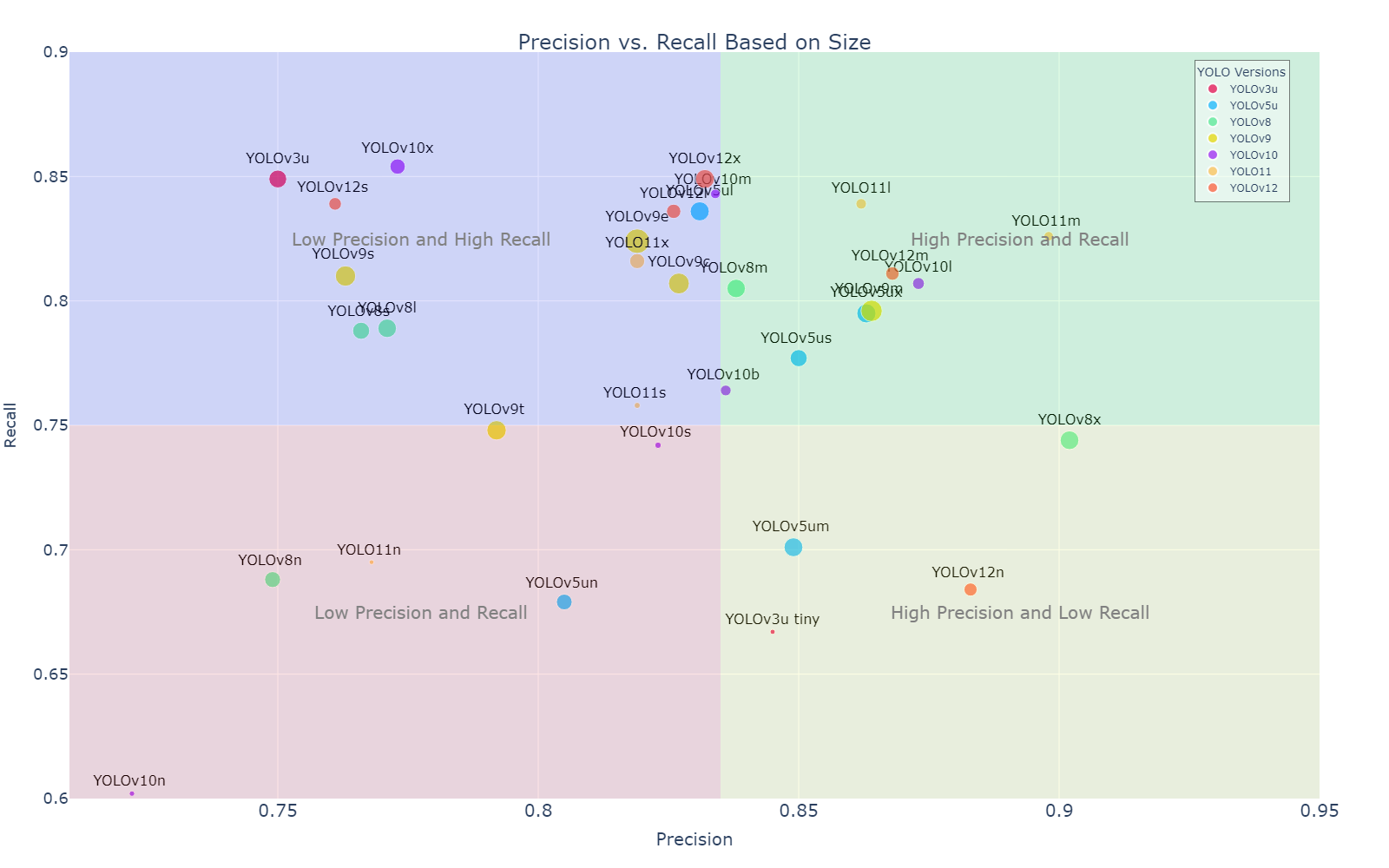}
    \caption{Precision vs. Recall based on size on traffic signs dataset, with larger circles indicating larger model sizes}
    \label{fig: prec signs}
\end{figure*}


 \begin{figure*}[htp]
    \centering
    \includegraphics[width=14cm, trim={0 0 0 2.5cm}, clip]{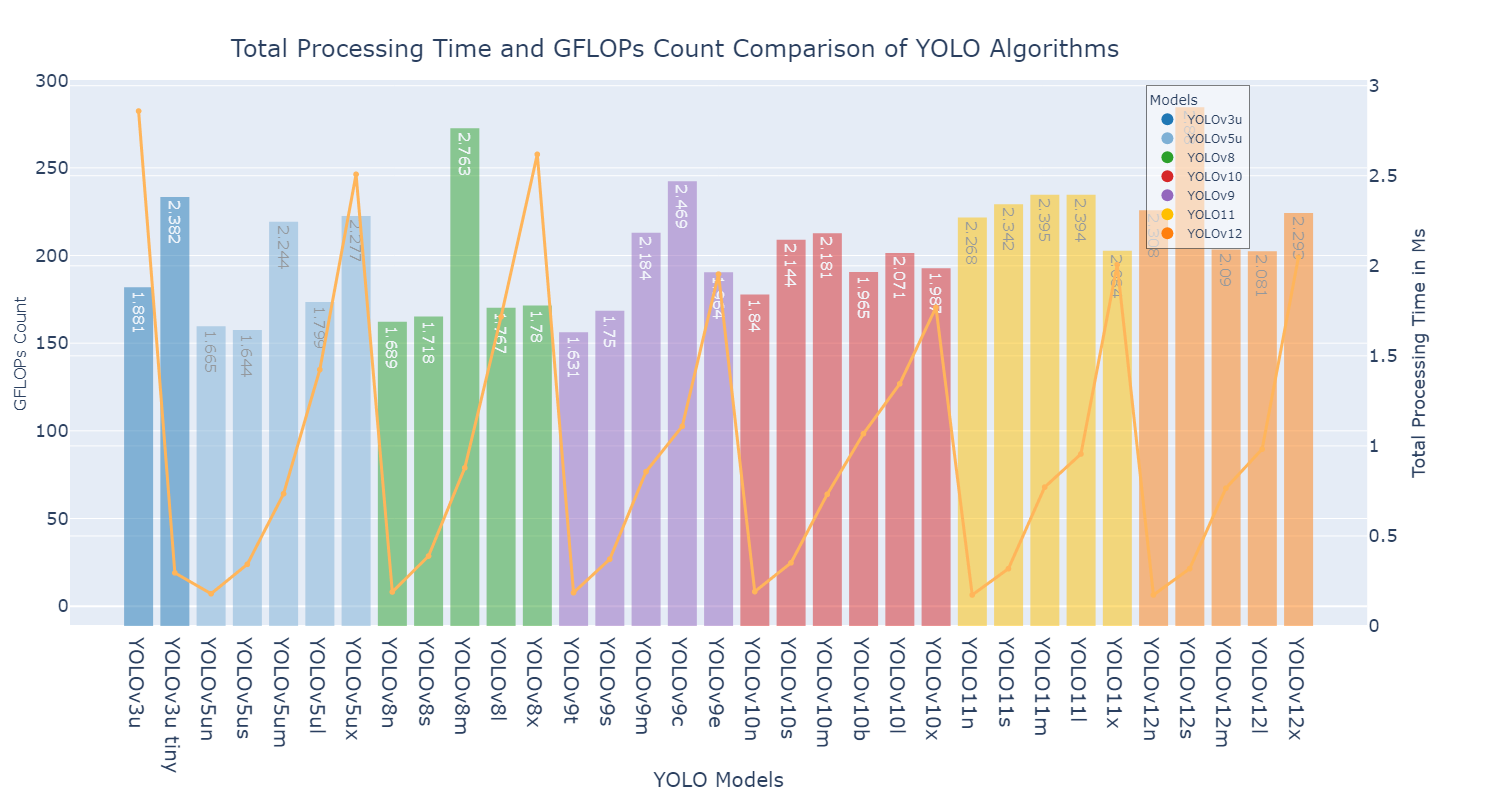}
    \caption{Total processing time and GFLOPs count results on traffic signs dataset}
    \vspace{-1em}
    \label{fig: time and gflop signs}
\end{figure*}
 
\paragraph{\textbf{Computational Efficiency:}}
Regarding computational efficiency, YOLOv10n is the most efficient, with a processing time of 2ms per image and a GFLOPs count of 8.3, as shown in Figure \ref{fig: time and gflop signs}. YOLO11n closely trails this at 2.2ms with a 6.4 GFLOPs count, and YOLOv3u tiny with a processing time of 2.4ms and a GFLOPs count of 19, making it relatively computationally inefficient compared to the other fast models. However, the data indicates that YOLOv9e, YOLOv9m, YOLOv9c, and YOLOv9s are the least efficient, with inference times of 16.1ms, 12.1ms, 11.6ms, and 11.1ms, and GFLOPs count of 189.4, 76.7, 102.6, and 26.8, respectively. These findings delineate a clear trade-off between accuracy and computational efficiency.

\paragraph{\textbf{Overall Performance:}}
When evaluating overall performance, which includes accuracy, size, and model efficiency, YOLO11m emerges as a consistently top-performing model as detailed in Figures \ref{fig: traffic signs table}, \ref{fig: map signs}, \ref{fig: prec signs}, and \ref{fig: time and gflop signs}. This is followed by YOLO11l and YOLOv10m showcasing the robustness of these models. On the contrary, YOLOv9e demonstrates poor performance overall in terms of accuracy, speed, GFLOPs, and model size. Additionally, YOLOv5um, YOLOv8m, and YOLOv8s performed suboptimally, further indicating the inconsistency in performance among earlier model families. Notably, the YOLO11 and YOLOv12 families significantly outperform other YOLO families regarding accuracy and computational efficiency. Their models consistently surpass counterparts from the YOLOv3u, YOLOv5u, YOLOv8, YOLOv9, and YOLOv10 families, demonstrating their ability to balance precision and speed effectively in the detection of objects with varying sizes.

\subsubsection{Africa Wildlife Dataset}
\begin{figure*}[htp]
    \centering
    \includegraphics[width=14cm]{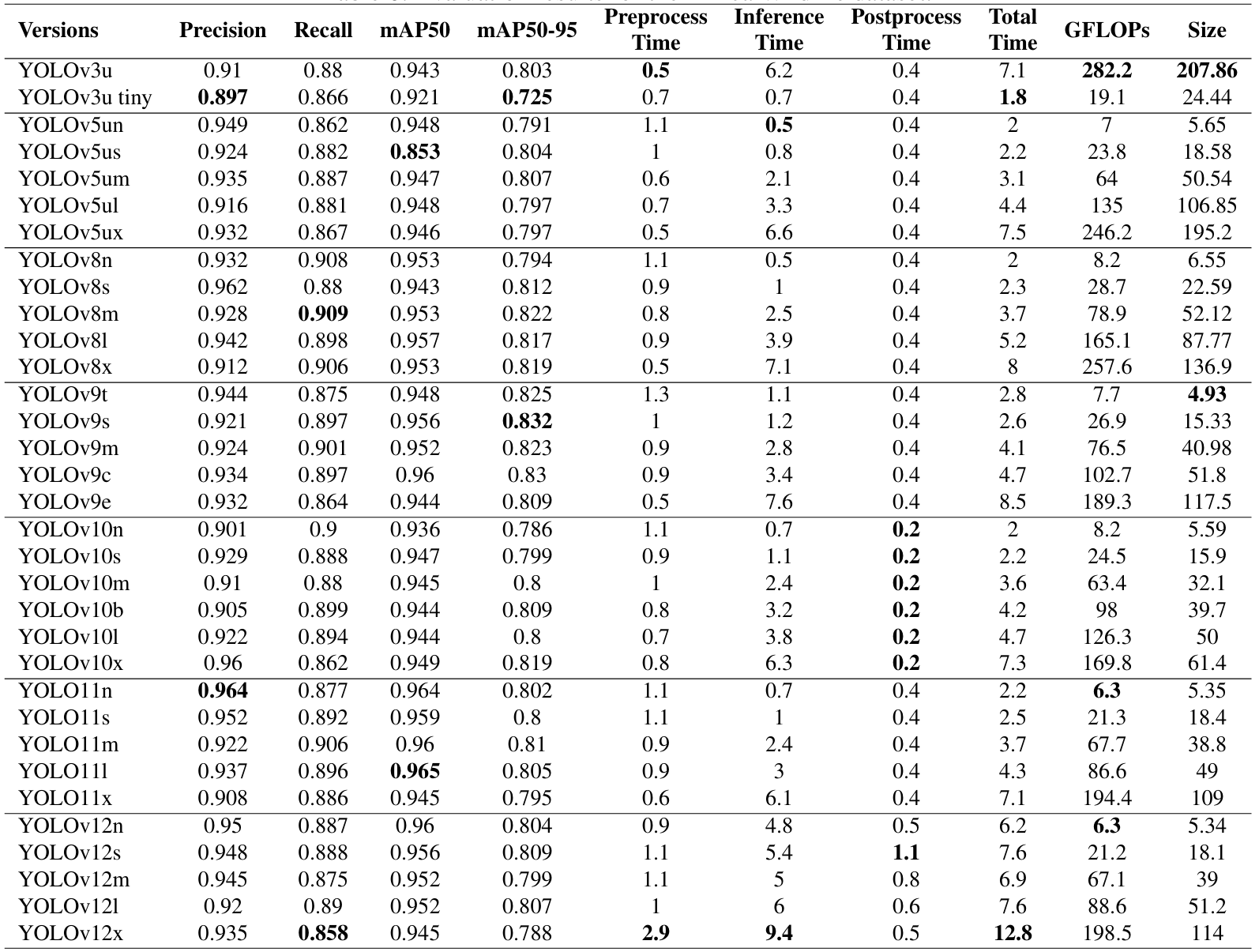}
    \caption{Evaluation results for the Africa wildlife dataset.}
    \label{fig:  africa table}
\end{figure*}
The results in Figure \ref{fig:  africa table} showcase the performance of the YOLO models on the Africa Wildlife dataset. This dataset contains large object sizes, focusing on the ability of YOLO models to predict large objects and their risk of overfitting due to the size of the dataset.

\begin{figure*}[htp]
    \centering
    \includegraphics[width=14cm, trim={0 0 0 1.7cm}, clip]{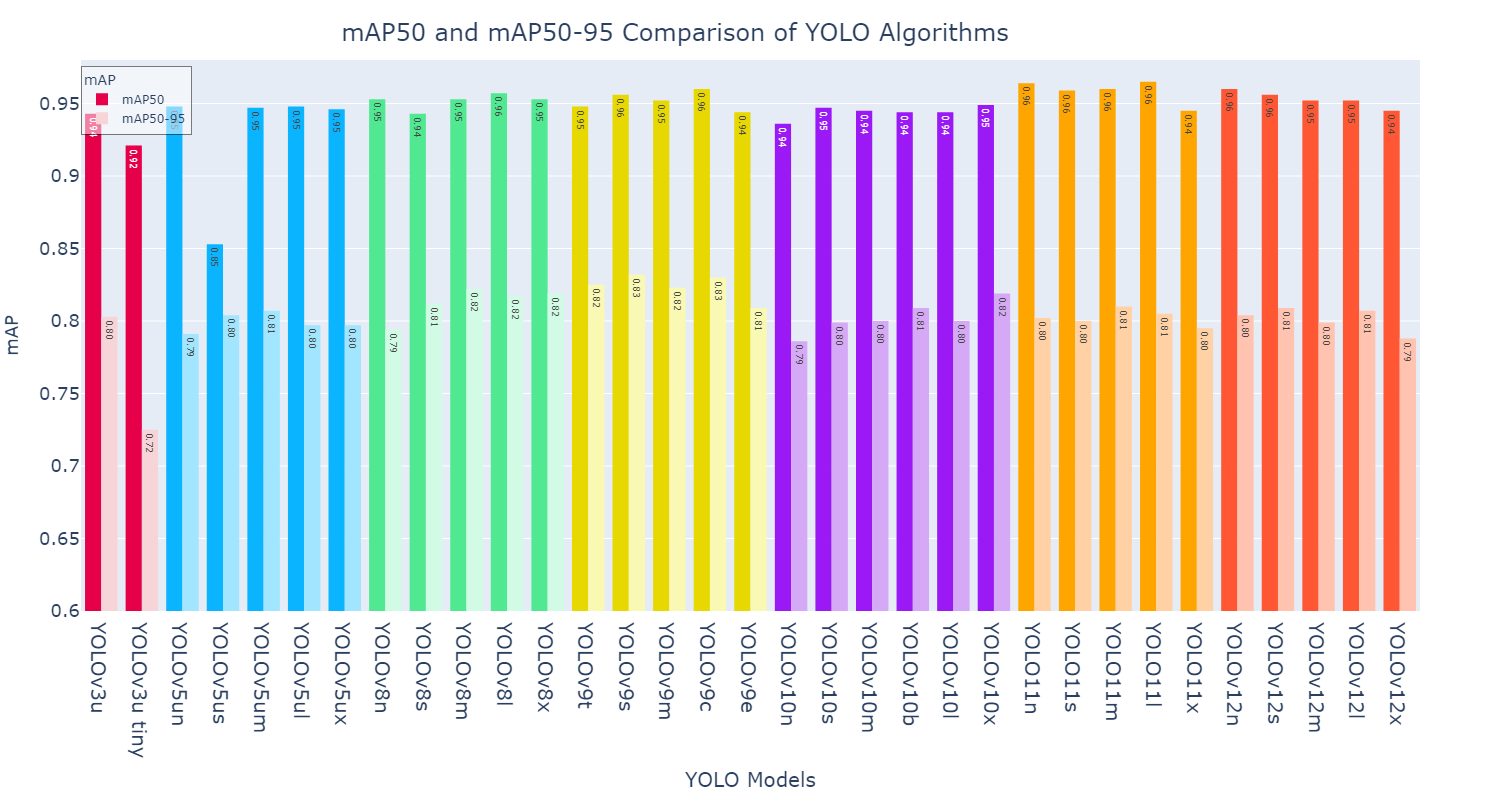}
    \caption{mAP50 and mAP50-95 YOLO results on Africa wildlife dataset. Each model is represented by two bars: the left bar shows the mAP50 score, while the right bar represents the mAP50-95 score}
    \label{fig: map africa}
\end{figure*}

\begin{figure*}[htp]
    \centering
    \includegraphics[width=14cm, trim={0 0 0 2cm}, clip]{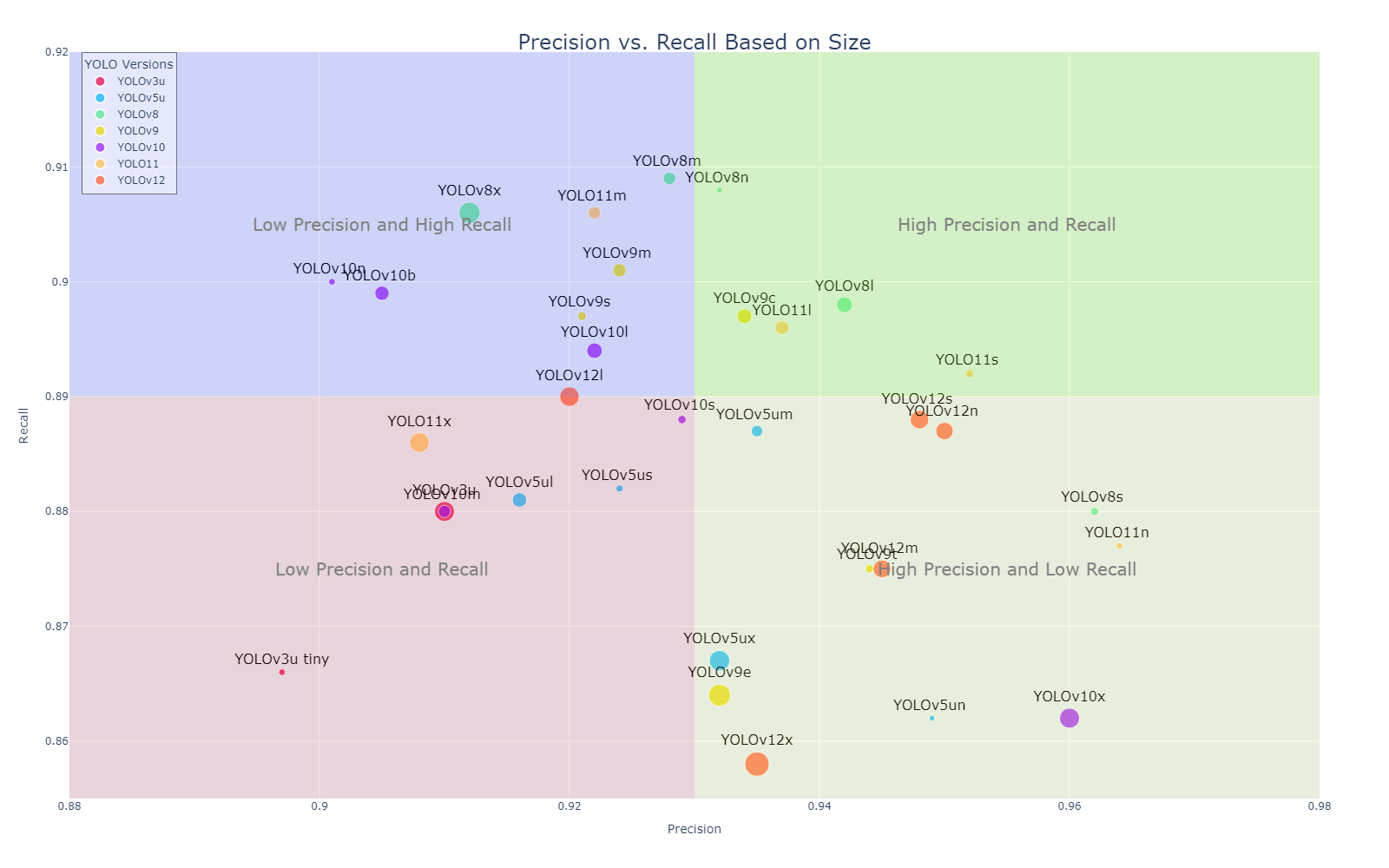}
    \caption{Precision vs. Recall based on size results on Africa wildlife dataset, with larger circles indicating larger model sizes}
    \label{fig: prec africa}
\end{figure*}

\begin{figure*}[htp]
    \centering
    \includegraphics[width=14cm, trim={0 0 0 2.5cm}, clip]{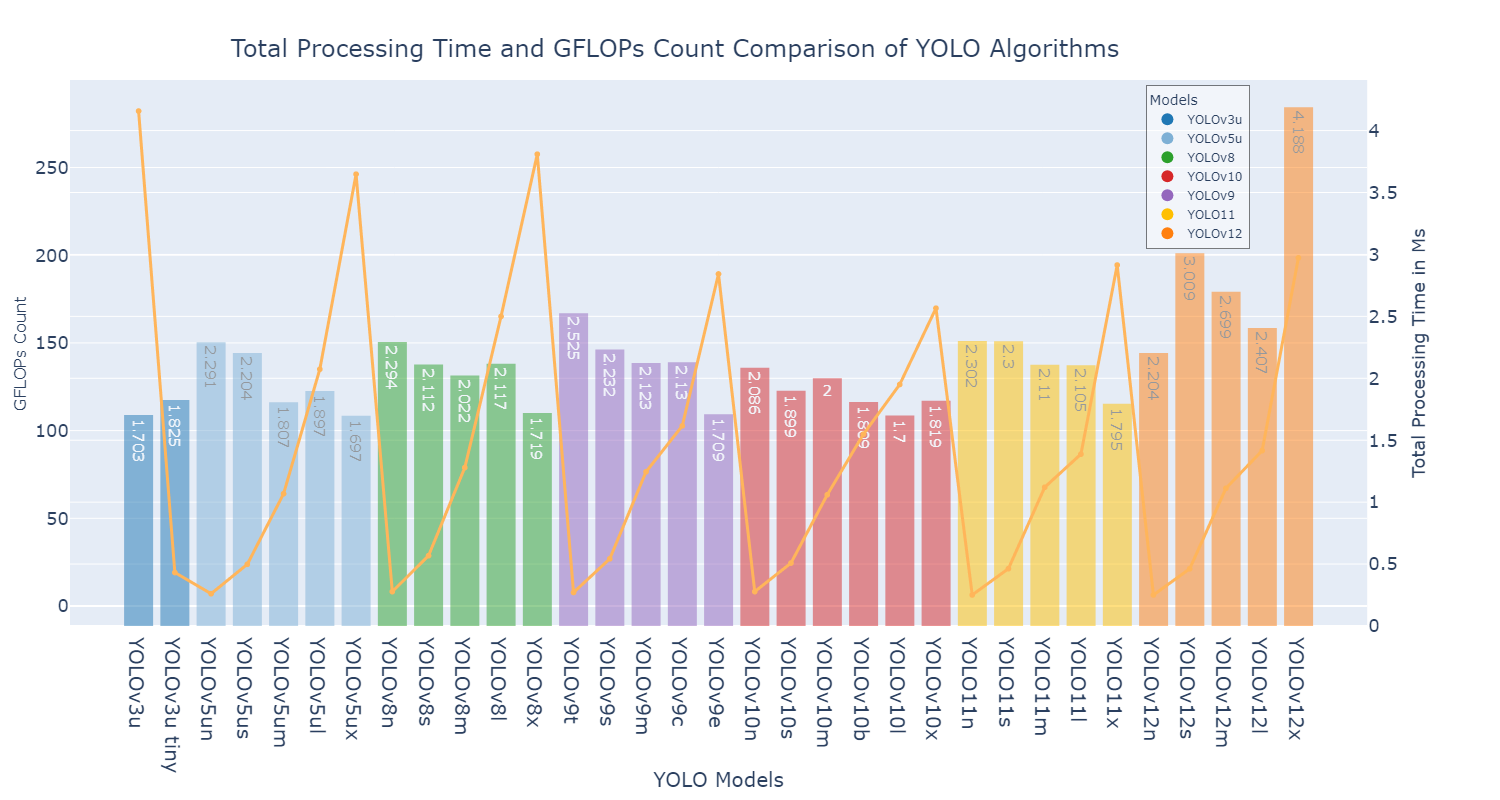}
    \caption{Total processing time and GFLOPs count results on Africa wildlife dataset}
    \vspace{-1em}
    \label{fig: time and gflop africa}
\end{figure*}

\paragraph{\textbf{Accuracy:}}
As illustrated in Figure \ref{fig: map africa}, YOLOv9s demonstrates exceptional performance with a high mAP50-95 of 0.832 and a mAP50 of 0.956, showcasing its robust accuracy across various IoU thresholds. YOLOv9c and YOLOv9t follow closely, with mAP50 scores of 0.96 and 0.948 and mAP50-95 scores of 0.83 and 0.825, respectively. These results highlight the YOLOv9 family’s ability to effectively learn patterns from a small sample of images, making it particularly suited for smaller datasets. In contrast, YOLOv5un, YOLOv10n, and YOLOv3u tiny show lower mAP50-95 scores of 0.791, 0.786, and 0.725, indicating their limitations in accuracy. The underperformance of larger models like YOLO11x, YOLOv5ux, YOLOv5ul, and YOLOv10l can be attributed to overfitting, especially given the small dataset size.

\paragraph{\textbf{Computational Efficiency:}}

 YOLOv10n, YOLOv8n, and YOLOv3u tiny are the fastest models, achieving processing times of 2ms and 1.8ms, with GFLOPs counts of 8.2 and 19.1, respectively. The first two models share the same processing speed and GFLOPs count, as showcased in Figure~\ref{fig: time and gflop africa}. Conversely, YOLOv12x exhibits the slowest processing time at 12.8 ms and a GFLOPs count of 198.5, followed by YOLOv9e at 8.5 ms and 189.3 GFLOPs count. These results indicate that larger models tend to require more processing time and hardware usage compared to smaller models, emphasizing the trade-off between model size and processing efficiency.

\paragraph{\textbf{Overall Performance:}}
YOLOv9t and YOLOv9s consistently excel across all metrics, delivering high accuracy while maintaining small model sizes, low GFLOPs, and short inference times, as shown in Figures \ref{fig:  africa table}, \ref{fig: prec africa}, \ref{fig: prec africa}, and \ref{fig: time and gflop africa}. This demonstrates the robustness of YOLOv9’s smaller models and their effectiveness on small datasets. In contrast, YOLOv12x and YOLO5ux show suboptimal accuracy despite their larger sizes and longer inference times, likely due to overfitting. Most large models underperformed on this dataset, with the exception of YOLOv10x, which benefited from a modern architecture that prevents overfitting.
\begin{figure*}
    \centering
    \includegraphics[width=14cm]{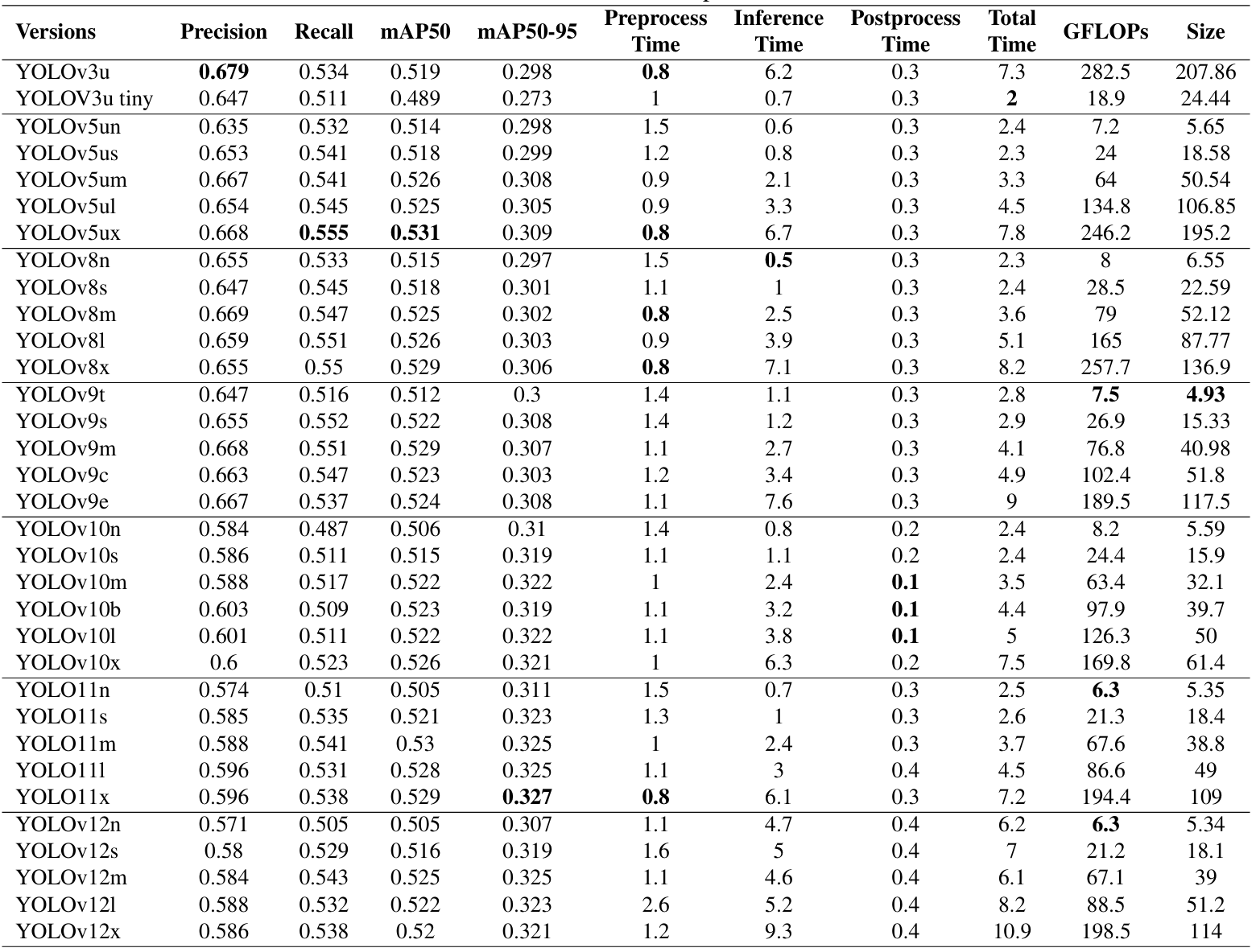}
    \caption{Evaluation results for the ships and vessels dataset}
    \label{fig: ships table}
\end{figure*}

\begin{figure*}
    \centering
    \includegraphics[width=14cm, trim={0 0 0 1.7cm}, clip]{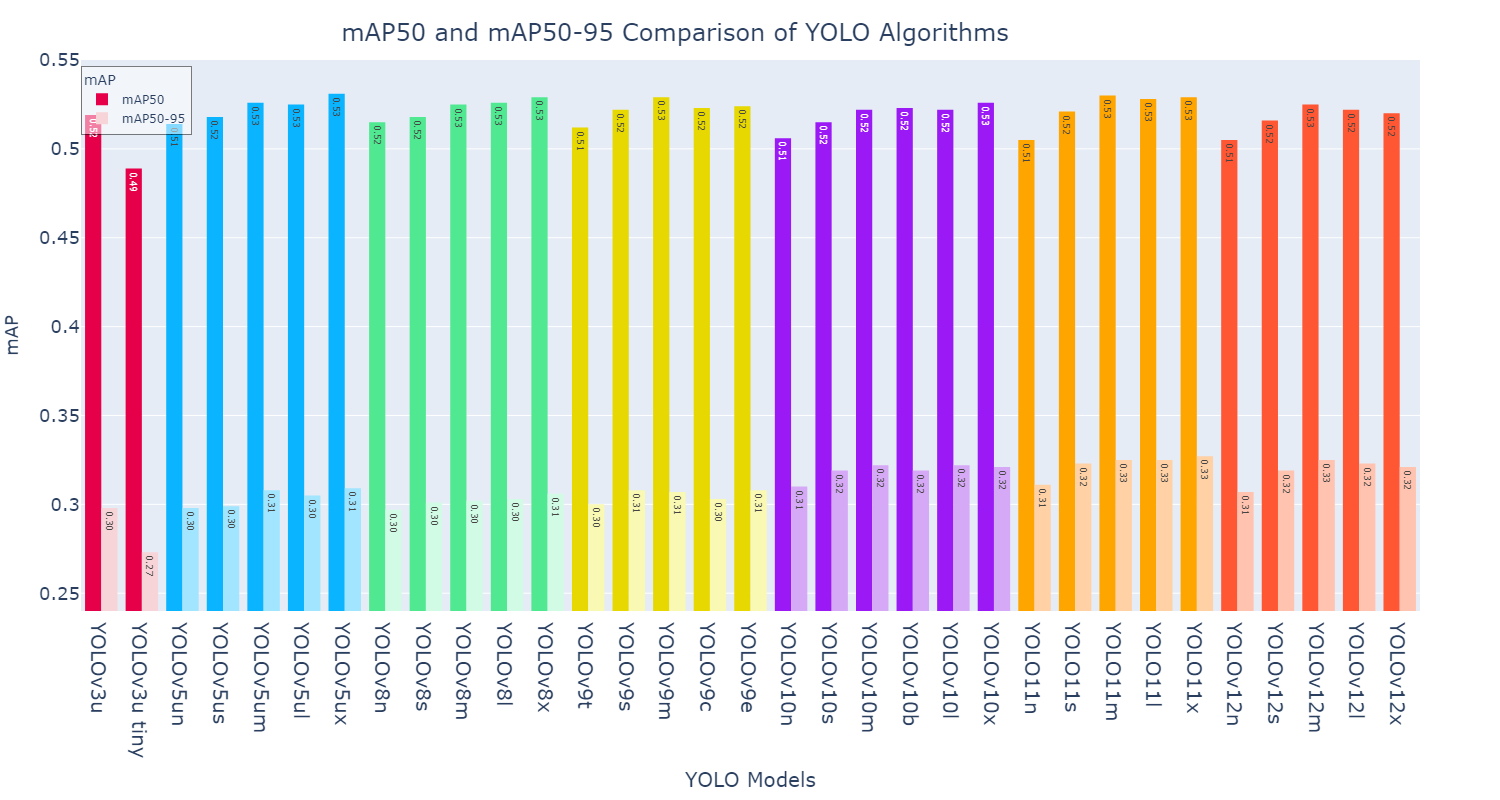}
    \caption{mAP50 and mAP50-95 YOLO results on ships and vessel dataset. Each model is represented by two bars: the left bar shows the mAP50 score, while the right bar represents the mAP50-95 score}
    \label{fig: map ships}
\end{figure*}

\begin{figure*}
    \centering
    \includegraphics[width=14cm, trim={0 0 0 2cm}, clip]{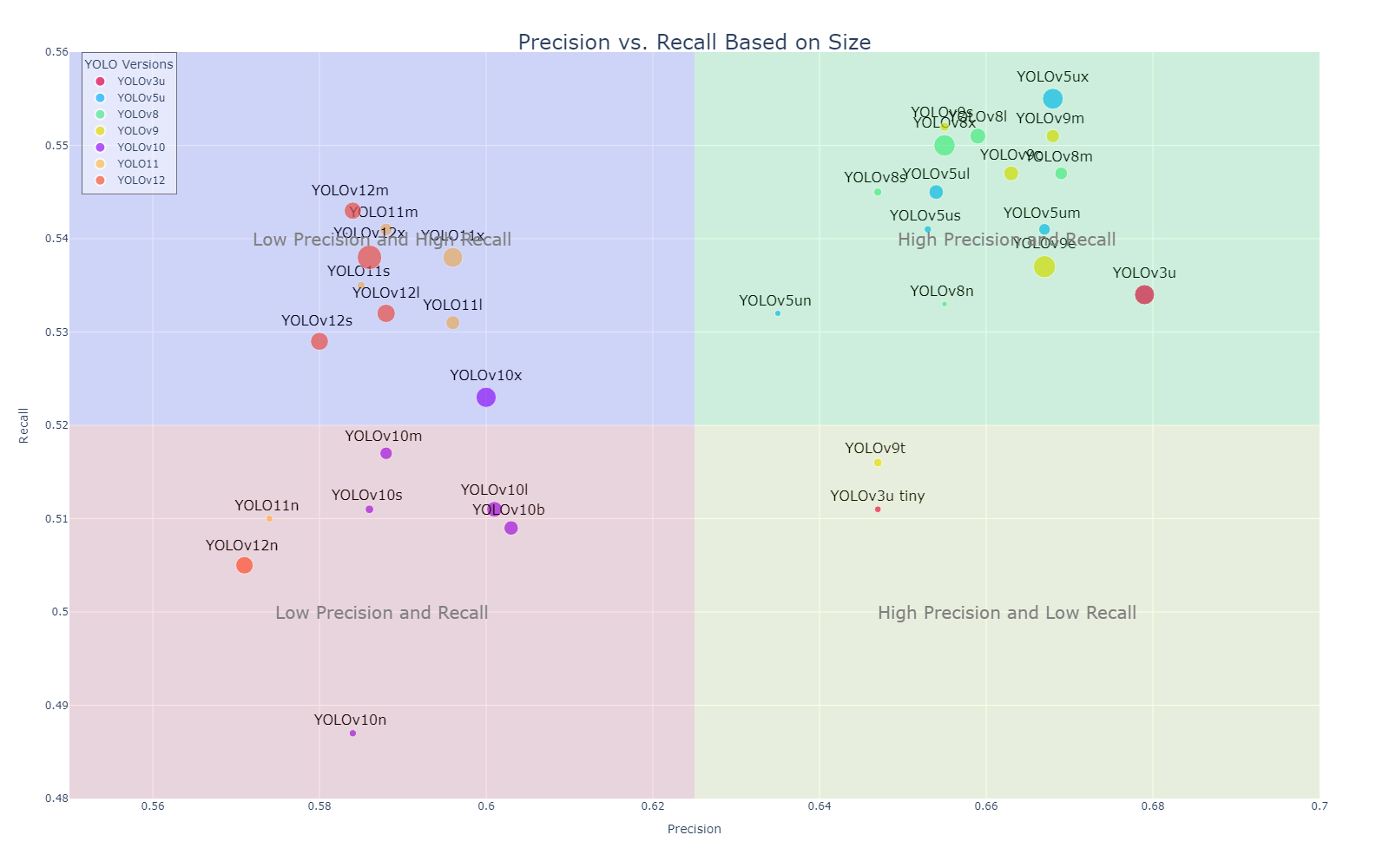}
    \caption{Precision vs. Recall based on size results on ships and vessels dataset, with larger circles indicating larger model sizes}
    \label{fig: prec ships}
\end{figure*}

\begin{figure*}
    \centering
    \includegraphics[width=14cm, trim={0 0 0 2.5cm}, clip]{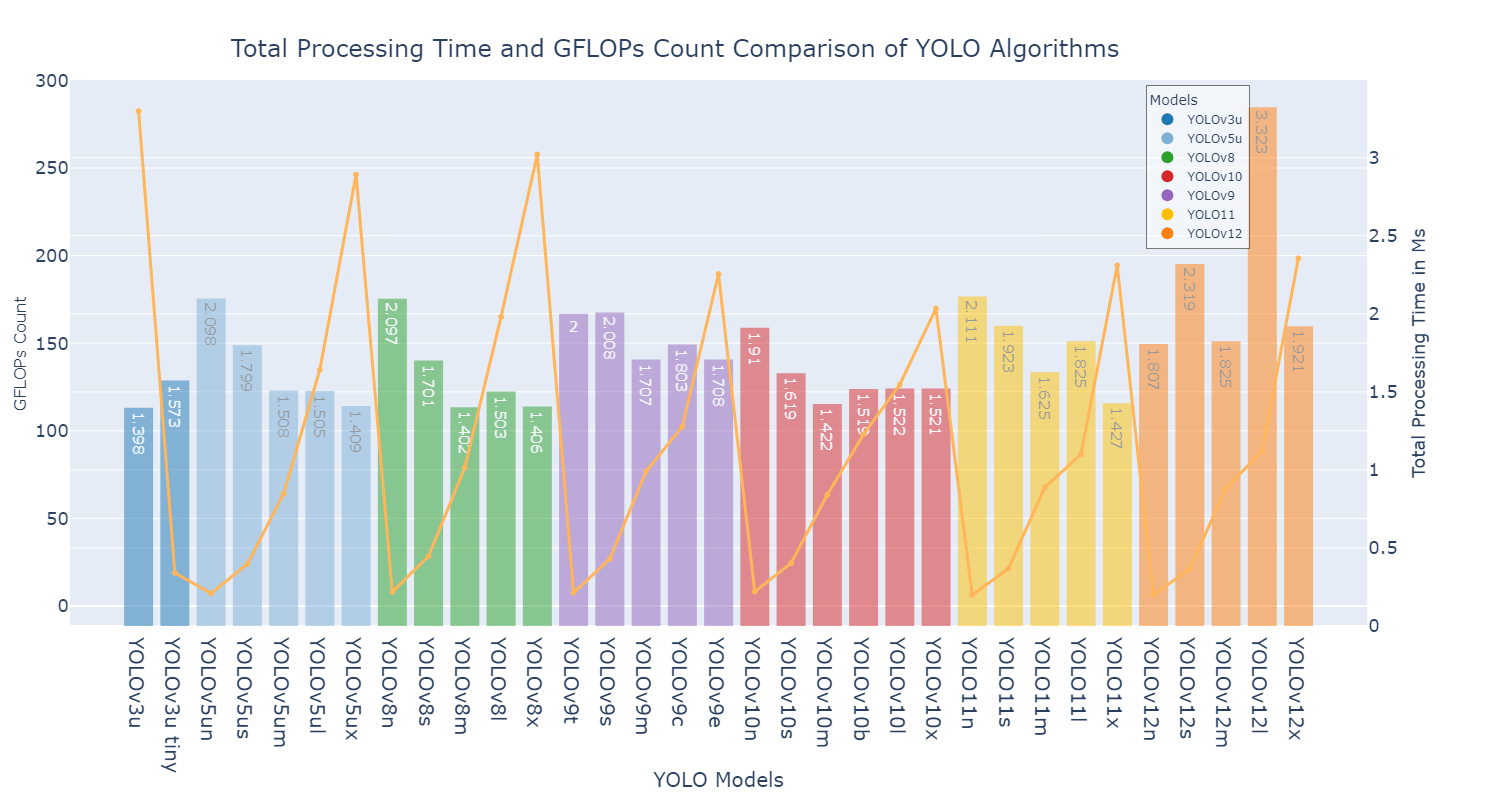}
    \caption{Total processing time and GFLOPs count results on ships and vessels dataset}
    \vspace{-1em}
    \label{fig: time and gflop ships}
\end{figure*}

\subsubsection{Ships and Vessels Dataset}
Figure \ref{fig: ships table}  presents the performance of YOLO models on the Ships and Vessels dataset, a large dataset featuring tiny objects with varying orientations.

\paragraph{\textbf{Accuracy:} }
The disparity between mAP50-95 and mAP50, illustrated in Figure \ref{fig: map ships}, underscores the challenges YOLO models face with higher IoU thresholds when detecting small objects. Additionally, YOLO models struggle with detecting objects of varying rotations. Among the models, YOLO11x achieved the highest accuracy, with a mAP50 of 0.529 and a mAP50-95 of 0.327, closely followed by YOLO11l, YOLO11m, and YOLO11s, which recorded mAP50 values of 0.529, 0.528, and 0.53, and mAP50-95 values of 0.327, 0.325, and 0.325, respectively. In contrast, YOLOv3u-tiny, YOLOv8n, YOLOv3u, and YOLOv5n exhibited the lowest accuracy, with mAP50 scores of 0.489, 0.515, 0.519, and 0.514, and mAP50-95 scores of 0.273, 0.297, 0.298, and 0.298, respectively. This suggests the outdated YOLOv3u architecture and the potential underfitting of smaller models with large datasets.

\paragraph{\textbf{Computational Efficiency:}}
As illustrated in Figure~\ref{fig: time and gflop ships}, YOLOv3u tiny achieved the fastest processing time at 2 ms, closely followed by YOLOv8n and YOLOv5un, recording 2.3 ms. YOLOv10 and YOLO11 models also excelled in speed, with YOLOv10n and YOLO11n achieving rapid inference times of 2.4 ms and 2.5 ms, along with GFLOPs counts of 8.2 and 6.3, respectively. In contrast, YOLOv12x exhibited the lowest speed, with 10.9 ms inference time and 198.5 GFLOPs count, highlighting the shortcomings of the newest YOLO family.

\paragraph{\textbf{Overall Performance:}}
The results in Figures \ref{fig: ships table}, \ref{fig: map ships}, \ref{fig: prec ships}, and \ref{fig: time and gflop ships} demonstrate that YOLO11s and YOLOv10s excelled in accuracy while maintaining compact sizes, low GFLOPs, and quick processing times. In contrast, YOLOv3u, YOLOv8x, and YOLOv8l fell short of expectations despite their larger sizes and longer processing times. These findings highlight the robustness and reliability of the YOLO11 family, particularly in improving the YOLO family's performance in detecting small and tiny objects while ensuring efficient processing. Additionally, the results reveal the underperformance of YOLOv9 models when faced with large datasets and small objects despite their modern architecture.

 \section{Discussion}

\begin{figure}
    \centering
    \includegraphics[width=7cm]{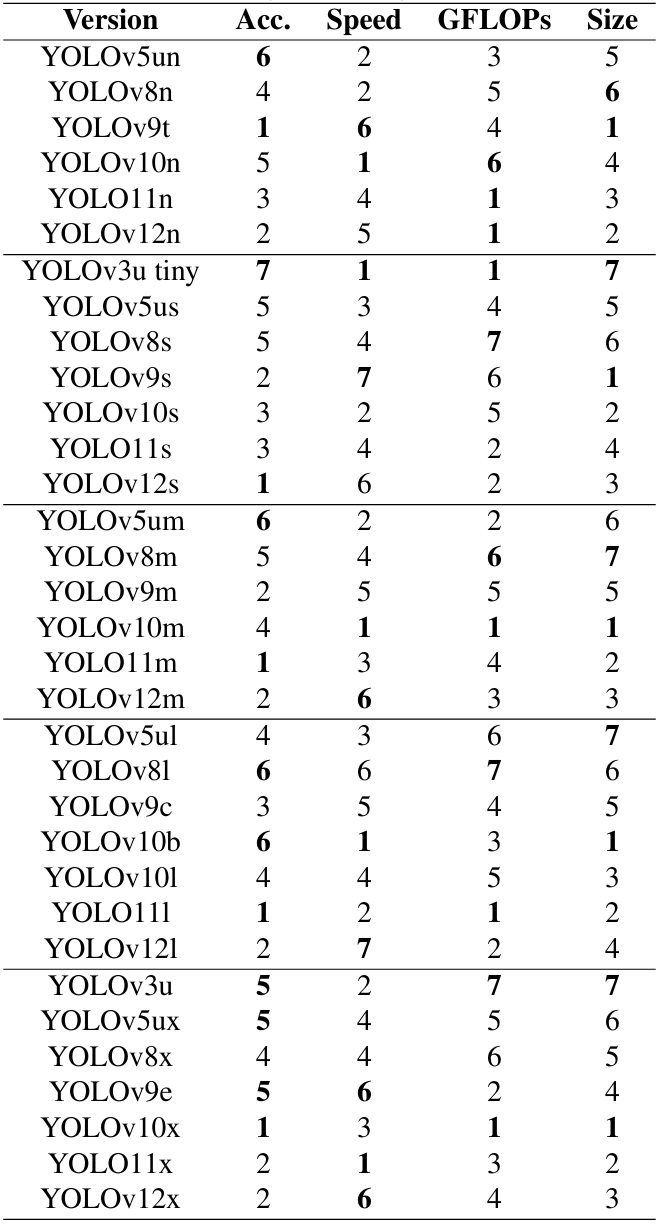}
    \caption{Overall ranking of YOLO algorithms based on size }
    \label{fig: overall rank}
\end{figure}

Models were ranked by accuracy, speed, GFLOPs count, and size across three datasets, as shown in Figure \ref{fig: overall rank}. To ensure fair comparisons, models were categorized by scale. YOLOv3u tiny was classified as small, YOLOv9t as nano, YOLOv9c as medium, YOLOv9c and YOLOv10b as large, and YOLOv9e as extra-large. Additionally, family-wide performance was analyzed by averaging rankings across all scales to compare overall performance as seen in Table \ref{tab:overall family rank}.

Accuracy was measured using mAP50-95 for a comprehensive performance assessment, while speed rankings were based on total processing time . Rankings range from Rank 1 (best) to the lowest performer, with results highlighted in bold. This approach ensures a fair evaluation of architectural improvements and computational trade-offs across model sizes and families.

\begin{table}[]
\caption{Overall ranking of YOLO algorithms based on family}

\label{tab:overall family rank}
\centering

\begin{tabular}{ccccc}
Version & \textbf{Acc.} & \textbf{Speed} & \textbf{GFLOPs} & \textbf{Size} \\ \hline
YOLOv3u & \textbf{7}    & \textbf{1}     & 4               & \textbf{7}    \\
YOLOv5u & 6             & 4              & 4               & 5             \\
YOLOv8  & 5             & 5              & \textbf{7}      & 6             \\
YOLOv9  & 3             & 6              & 6               & 4             \\
YOLOv10 & 4             & 2              & 3               & \textbf{1}    \\
YOLO11  & \textbf{1}             & 3              & \textbf{1}      & 2             \\
YOLOv12 & \textbf{1}    & \textbf{7}     & 2               & 3             \\ \hline
\end{tabular}
\end{table}

\subsection{Performance Analysis by Model Scale}

\paragraph{Nano Models}
Among nano models, YOLOv9t ranked highest in accuracy and speed, benefiting from its lightweight GELAN architecture, which optimizes parameter efficiency while maintaining strong detection capabilities. However, YOLOv10n, despite its computational efficiency, exhibited lower accuracy due to its reliance on the One-to-One head and NMS-Free training, which struggled with densely packed objects.

\paragraph{Small Models}
YOLOv12s emerged as the best small model in terms of accuracy, likely due to its R-ELAN-based feature aggregation. However, it performed poorly in speed, as the added complexity of Area Attention (A2) increased inference time. In contrast, YOLOv10s and YOLO11s provided a balanced performance, with YOLO11s leveraging the C3k2 block for improved feature extraction and YOLOv10s excelling in speed due to its efficient head design.

\paragraph{Medium Models}
YOLO11m achieved the best balance of accuracy and efficiency in this category, owing to C2PSA, which enhances spatial feature capture. YOLOv10m outperformed all models in terms of computational efficiency but had lower accuracy due to its reliance on NMS-Free training. Meanwhile, YOLOv9m, while achieving strong accuracy, ranked lower in efficiency due to its reliance on gradient-based optimizations, which added computational overhead.

\paragraph{Large Models}
YOLO11l outperformed other large models in accuracy and speed, thanks to the integration of C2PSA and refined convolutional operations. In contrast, YOLOv9c and YOLOv12l struggled with efficiency, with YOLOv9c being computationally expensive and YOLOv12l suffering from latency due to FlashAttention’s memory overhead.

\paragraph{Extra-Large Models}
YOLO11x provided the best balance of accuracy and computational efficiency, with its C3k2-based optimization reducing inference overhead. However, YOLOv9e and YOLOv12x exhibited slower speeds and inefficient GFLOPs utilization, highlighting the limitations of their respective architectures when scaling up.

\subsection{Performance Analysis by Family}
\paragraph{YOLOv12}
YOLOv12 delivered strong accuracy results, ranking first alongside YOLO11 in accuracy across models. This performance can be attributed to its integration of the Area Attention Module (A2) and Residual Efficient Layer Aggregation Networks (R-ELAN), which enhanced feature extraction and contextual understanding. However, this came at the cost of computational efficiency, as the added complexity introduced significant latency and reduced processing speed. FlashAttention, while designed to optimize memory access, further contributed to increased inference time. While YOLOv12 excelled in accuracy, its slow processing speed and higher computational demands limited its overall practicality, highlighting the trade-off between architectural complexity and real-world efficiency.
\paragraph{YOLO11}
The YOLO11 family consistently ranked among the best due to its C3k2 block and C2PSA, enhancing efficiency and contextual understanding by ranking first in terms of accuracy and GFLOPs count and among the first in terms of speed and size. These innovations enabled YOLO11 to achieve high accuracy with low computational overhead, making it a well-balanced choice. While YOLOv8 was widely used for various tasks such as Pose Estimation and OBB, YOLO11 has emerged as a superior alternative, offering better feature extraction, inference speed, and accuracy. This study establishes YOLO11 as a new benchmark, demonstrating its high detection accuracy and low-latency processing, and various capabilities, making it ideal for real-world applications requiring both speed and precision.

\paragraph{YOLOv10}
YOLOv10 models were highly efficient, particularly in speed and size, due to the One-to-One head and NMS-Free training. However, their accuracy was subpar compared to YOLOv9, YOLO11, and YOLOv12, particularly in scenarios requiring precise bounding box refinement, such as the Traffic Signs dataset.

\paragraph{YOLOv9}
Despite strong accuracy due to PGI and GELAN, YOLOv9 models suffered from slow inference times, limiting their real-time applicability. Their architectural choices prioritized precision over efficiency, making them suitable for applications where accuracy is paramount but speed is less critical.

\paragraph{YOLOv8, YOLOv5u, and YOLOv3u}
While YOLOv8, YOLOv5usurpassed YOLOv3u in accuracy, they lagged behind newer models in overall efficiency. YOLOv8 introduced the C2k module for improved feature extraction, but its higher computational cost made it less efficient than more recent architectures. With YOLO11 offering superior accuracy, speed, and efficiency, YOLOv8 is no longer the go-to model for real-world applications, as YOLO11 provides a better balance between performance and computational overhead. Meanwhile, YOLOv5u improved on YOLOv3u but lacked modern enhancements like self-attention, limiting its scalability.

Overall, the rankings highlight the dominance of YOLO11 as the most balanced family, with YOLOv10 excelling in efficiency and YOLOv9 maintaining strong accuracy at the cost of speed. YOLOv12, despite its ambitious design and exceptional accuracy, failed to outperform its predecessors due to increased latency, reinforcing the importance of balancing architectural complexity with real-world feasibility.

\subsection{Dataset Size}
The size of the dataset significantly influences the performance of YOLO models. For instance, large models did not perform optimally on the small African wildlife dataset compared to their results on the Traffic Signs and Ships and Vessels datasets due to being more prone to overfitting. Conversely, small models like YOLOv9t and YOLOv9s performed best on the Africa Wildlife dataset, showcasing the effectiveness of small-scaled models when handling limited datasets.

\subsection{Impact of Training Datasets}
The performance of YOLO models is influenced by the training datasets used, as shown in Figures \ref{fig: traffic signs table}, \ref{fig:  africa table}, and \ref{fig: ships table}. Different datasets yield varying results and top performers, indicating that dataset complexity affects algorithm performance. This underscores the importance of using diverse datasets during benchmarking to obtain comprehensive results on the strengths and limitations of each model. Additionally, this highlights the need for a balanced consideration of accuracy, speed, and model size when selecting YOLO models for specific applications.

\section{Real-Life Applications} \label{real-life applications}

\begin{table*}[h!]
\centering
\caption{YOLO Model Recommendations for Real-Life Applications}

\resizebox{\textwidth}{!}{
\begin{tabular}{lll}
\hline
\textbf{Scenario}                                                                                    & \textbf{Recommended YOLO Models} & \textbf{Application Examples}                                                                                                \\ \hline
\multirow{2}{*}{\begin{tabular}[c]{@{}l@{}}Computationally Constrained \\ Environments\end{tabular}} & YOLOv10(n/s/m)                   & \multirow{2}{*}{\begin{tabular}[c]{@{}l@{}}Drone surveillance, embedded \\ systems, battery-powered devices \cite{yolov10eff, yolov10eff2, yolov10eff3, yolo11eff, yolo11eff2, yolo11eff3} \end{tabular}}    \\
                                                                                                     & YOLO11(n/s/m)                    &                                                                                                                              \\ \hline
\multirow{2}{*}{\begin{tabular}[c]{@{}l@{}}Real-Time Monitoring \\ and Rapid Response\end{tabular}}  & YOLOv10(n/s/m)                   & \multirow{2}{*}{\begin{tabular}[c]{@{}l@{}}Autonomous vehicles, disaster \\ response, traffic monitoring \cite{yolov10eff, yolov10eff2, yolov10eff3, yolo11effrealtime, yolo11eff, yolo11eff2, yolo11eff3} \end{tabular}}       \\
                                                                                                     & YOLO11(n/s/m)                    &                                                                                                                              \\ \hline
\multirow{2}{*}{Detection of Small Objects}                                                          & YOLO11(m/l/x)                    & \multirow{2}{*}{\begin{tabular}[c]{@{}l@{}}Wildlife tracking, satellite imaging \\ of ships and small vehicles \cite{yolo11effrealtime, yolo11eff, yolo11eff2, yolo11eff3} \end{tabular}} \\
                                                                                                     & YOLOv12(m/l/x)                   &                                                                                                                              \\ \hline
\multirow{2}{*}{Detection of Large Objects}                                                          & YOLOv8(s/m/l/x)                  & \multirow{2}{*}{\begin{tabular}[c]{@{}l@{}}Satellite imaging \\ (buildings, forests, urban development)  \cite{yolov9largeimg, yolov8largeimg, yolov8largeimg2}\end{tabular}}        \\
                                                                                                     & YOLOv9(t/s/m/c/e)                &                                                                                                                              \\ \hline
\begin{tabular}[c]{@{}l@{}}Overlapping and Densely \\ Packed Objects\end{tabular}                    & YOLOv9(t/s/m/c/e)                & \begin{tabular}[c]{@{}l@{}}Traffic intersections, \\ wildlife in dense foliage  \end{tabular}                                  \\ \hline
\multirow{2}{*}{\begin{tabular}[c]{@{}l@{}}Oriented and Rotated \\ Object Detection\end{tabular}}    & YOLO11 (OBB Support)             & \multirow{2}{*}{\begin{tabular}[c]{@{}l@{}}Satellite imagery, \\ aerial monitoring of ships  \cite{yolov9overlap, yolov9overlap2}\end{tabular}}                    \\
                                                                                                     & YOLOv12 (OBB Support)            &                                                                                                                              \\ \hline
\multirow{2}{*}{Handling Large Datasets}                                                             & YOLO11(m/l/x)                    & \multirow{2}{*}{\begin{tabular}[c]{@{}l@{}}City-wide monitoring, \\ biodiversity mapping \cite{yolo11large, yolo11large2} \end{tabular}}                       \\
                                                                                                     & YOLOv12(m/l/x)                   &                                                                                                                              \\ \hline
Handling Small Datasets                                                                              & YOLOv9(t/s/m/c)                  & \begin{tabular}[c]{@{}l@{}}Anti-poaching monitoring, \\ niche industrial tasks \cite{yolov9smalldata, yolov9small, yolov9small2} \end{tabular}                                  \\ \hline
\end{tabular}}
\label{tab:yolo_applications}
\end{table*} 

The advancements in YOLO models have enabled their application across diverse real-world scenarios, each with distinct demands for model performance, such as computational efficiency, accuracy, and adaptability to varying object complexities. Based on the benchmark of various YOLO models and related papers, this section highlights how different models excel under specific scenarios, making them suitable for deployment in various tasks across industries, as shown in Table \ref{tab:yolo_applications}.

\subsection{Computationally Constrained Environments}  
In scenarios with limited computational resources, such as drones, embedded systems, and battery-powered devices, models must be lightweight and efficient to avoid draining power or overloading hardware. The nano, small, and medium-sized models of the YOLO11 and YOLOv10 families are optimal choices due to their high accuracy, compact model sizes, low inference times, and low GFLOPs count leading to minimal resource consumption, as also seen in related papers \cite{yolov10eff, yolov10eff2, yolov10eff3, yolo11eff, yolo11eff2, yolo11eff3}.

\subsection{Real-Time Monitoring and Rapid Response}  
For applications requiring real-time detection, such as autonomous driving, surveillance, and disaster response, models need to balance accuracy and speed. As illustrated by this study and related research papers \cite{yolov10eff, yolov10eff2, yolov10eff3, yolo11effrealtime, yolo11eff, yolo11eff2, yolo11eff3}, the nano, small, and medium-sized YOLO11 and YOLOv10 models demonstrate strong performance in maintaining fast inference while accurately detecting objects across various scales.

\subsection{Detection of Small and Large Objects}  
Object detection models often face the challenge of accurately identifying small and large objects. Small objects, such as wildlife or distant vehicles, require models capable of capturing fine spatial details. Medium, large, and extra-large-sized YOLO11 and YOLOv12 models excel in these scenarios, as illustrated in our paper and related work \cite{yolo11effrealtime, yolo11eff, yolo11eff2, yolo11eff3}. Meanwhile, all YOLOv9 and YOLOv8 models offer robust performance for detecting large objects, making them ideal for scenarios such as satellite imaging of buildings, forests, and urban development \cite{yolov9largeimg, yolov8largeimg, yolov8largeimg2}.

\subsection{Overlapping and Densely Packed Objects}  
Detecting objects that overlap or appear densely packed is a major challenge in traffic intersections or wildlife monitoring scenarios. In these cases, models must efficiently distinguish between objects and avoid false positives. All YOLOv9 models are highly effective in handling overlapping objects, as seen in this benchmark and related work \cite{yolov9overlap, yolov9overlap2}. This suits them, particularly for applications like detecting animals in dense foliage or distinguishing closely packed traffic signs.

\subsection{Oriented Object Detection}  
Objects often appear at varying angles or orientations in satellite imaging and certain surveillance applications, requiring models with specialized detection capabilities. The YOLO11 and YOLOv12 families' support for OBB enables effective detection of rotated objects, achieving higher accuracy than YOLOv8 models.

\subsection{Handling Large and Very Small Datasets}  
    Large-scale datasets with diverse object categories require models that generalize well without overfitting. Medium, large, and extra-large YOLO11 and YOLOv12 models excel in such scenarios, as showcased in our paper and related studies \cite{yolo11large, yolo11large2}. For tiny datasets, where overfitting is a concern, all YOLOv9 models are optimal choices. Their compact sizes and efficient parameter usage enable effective learning from limited data \cite{yolov9smalldata, yolov9small, yolov9small2}.

\section{Conclusion}
\label{sect:conclusion}
This benchmark study thoroughly evaluates the performance of various YOLO algorithms. It pioneers a comprehensive comparison of YOLOv12 against its predecessors and assesses their performance across three diverse datasets: Traffic Signs, African Wildlife, and Ships and Vessels. The datasets were carefully selected to encompass various object properties, including varying object sizes, aspect ratios, and object densities. We showcase the strengths and weaknesses of each YOLO version and family by examining a wide range of metrics such as Precision, Recall, Mean Average Precision (mAP), Processing Time, GFLOPs count, and Model Size. Our study addresses the following key research questions:
\begin{itemize}
    \item Which YOLO algorithm demonstrates superior performance across a comprehensive set of metrics?
    \item How do different YOLO versions perform on datasets with diverse object characteristics?
    \item What are each YOLO version's specific strengths and limitations, and how can these insights inform the selection of suitable algorithms for various applications?
    \item Which YOLO models are best suited for specific real-world scenarios?
\end{itemize}

In particular, the YOLO11 family emerged as the most consistent, with YOLO11m striking an optimal balance between accuracy, efficiency, and model size. While YOLOv10 delivered slightly lower accuracy than YOLO11, it excelled in speed and efficiency, making it a strong choice for applications requiring efficiency and fast processing. Additionally, YOLOv9 performed well overall and particularly stood out in smaller datasets. These findings provide valuable insights for industry and academia, guiding the selection of the most suitable YOLO algorithms and informing future developments and enhancements.

Despite being the newest addition to the YOLO family, YOLOv12 delivered an underwhelming performance. Its ambitious architectural changes, including the Area Attention Module and R-ELAN, introduced complexity without translating into clear advantages. The model struggled to balance accuracy and speed, highlighting the challenges of effectively integrating advanced attention mechanisms within the YOLO framework.

While the evaluated algorithms show promising performance, there is room for improvement. Future research should focus on optimizing YOLOv10’s accuracy while maintaining its speed and efficiency. Additionally, simplifying YOLOv12’s complex attention mechanisms could enhance its speed without compromising accuracy. Further architectural advancements could lead to even more efficient and accurate YOLO models. Our future work will address these gaps and explore enhancements to maximize overall efficiency and real-world applicability.

 \section*{Declarations}
\begin{itemize}
\item Funding: Not applicable
\item Conflict of interest/Competing interests: The authors declare no conflict of interest.
\item Ethics approval and consent to participate: Not applicable
\item Consent for publication: The authors confirm that all necessary permissions have been obtained for the publication of this work, including the use of copyrighted material and any personal information or sensitive data.
\item Data availability: Data is available on GitHub and will be publicly available upon publication of the final version 
\item Materials availability: Not applicable
\item Code availability: Source code is available on GitHub and will be publicly available upon publication of the final version.
\item Author contribution: All authors contributed to the study conception and design. Material preparation, data collection, and analysis were performed by Nidhal Jegham and Chan Young Koh. The first draft of the manuscript was written by Nidhal Jegham, and all authors commented on and edited previous versions of the manuscript. All authors read and approved the final manuscript. Dr. Abdeltawab Hendawi supervised the work of this study.

\end{itemize}

\bibliographystyle{plain}

\bibliography{sn-bibliography}

\end{document}